\documentclass{article}

\usepackage{PRIMEarxiv}

\usepackage[utf8]{inputenc} 
\usepackage[T1]{fontenc}    
\usepackage[colorlinks=true,linkcolor=blue!50!black,citecolor=blue!50!black,urlcolor=blue!50!black]{hyperref}       
\usepackage{url}            
\usepackage{xcolor}
\usepackage{booktabs}       
\usepackage{amsfonts}       
\usepackage{amsmath,amssymb}
\usepackage{nicefrac}       
\usepackage{microtype}      
\usepackage{fancyhdr}       
\usepackage{graphicx}       
\usepackage{subcaption}
\usepackage{float}
\graphicspath{{media/}}     

\title{HQNN-CL: Regime-Aware Hybrid Quantum Neural Network with Continual Learning for Financial Forecasting}

\author{
  Muhammad Atif Saeed\thanks{Corresponding author.} \\
  Department of Artificial Intelligence \& Data Science \\
  National University of Computer \& Emerging Sciences \\
  Islamabad, Pakistan \\
  \texttt{atif.saeed@isb.nu.edu.pk} \\
  \And
  Akhtar Jamil \\
  Department of Computer Science \\
  National University of Computer \& Emerging Sciences \\
  Islamabad, Pakistan \\
  \texttt{akhtar.jamil@nu.edu.pk} \\
  \And
  Subhan Ullah \\
  Department of Cyber Security \\
  National University of Computer \& Emerging Sciences \\
  Islamabad, Pakistan \\
  \texttt{subhan.ullah@nu.edu.pk} \\
}

\begin{document}
\maketitle

\begin{abstract}
Financial time-series forecasting is still difficult owing to the non-stationary nature of the market, regime changes, and catastrophic forgetting, which makes conventional deep learning models hard to adapt in the long term. To achieve uncertainty-aware robust financial forecasting in the face of changing market conditions, this study proposes a Hybrid Quantum-Classical Neural Network with Continual Learning (HQNN-CL). The proposed framework combines Gaussian Hidden Markov Model (HMM) based regime detection, BiLSTM temporal encoding, adaptive-depth parameterized quantum circuits, Elastic Weight Consolidation (EWC), experience replay, and probabilistic forecasting in a single architecture. The framework was tested on four real-time stock datasets: Microsoft, Systems Limited, Teradata and Tata Consultancy Services, and compared with seven state-of-the-art forecasting models, including recurrent and Transformer-based models. The experimental results show that HQNN-CL has outperformed other methods on the Systems Limited (RMSE = 3.0149) and Teradata (RMSE = 0.8909) datasets and is within 0.32\% and 0.40\% of the best baseline method for Microsoft and Tata Consultancy Services, respectively. The proposed framework also generates well-calibrated probabilistic forecasts within the range of 89.5\% to 96.5\% empirical coverage rate, has zero catastrophic forgetting (BWT = 0.0000) for all sequential learning tasks, and consistently shows positive forward transfer. Ablation and statistical significance analyses also validate the performance of adaptive quantum feature learning and continual adaptation under heterogeneous market conditions, regime-wise. In addition to competitive forecasting accuracy, HQNN-CL offers tangible advantages, such as uncertainty-aware risk assessment, regime-adaptive learning, and fewer retraining needs, fitting to the continuously changing landscape of finance. The results show the hybrid quantum learning with continual adaptation offers a practical and scalable framework for the next-generation intelligent financial forecasting.
\end{abstract}


\section{Introduction}
In quantitative finance, predicting future trends in financial time-series is one of the basic problems, as accurate forecasts can aid in portfolio management, risk assessment, and making investment decisions\cite{P6}. The difficulty arises in that financial markets are naturally non-stationary and characterized by sudden shifts from a low volatility regime to a transitional regime and to a high volatility regime~\cite{P16}. The statistical relationships learnt from the historical observations could thus lose its information as market conditions change. Furthermore, models that are used for a long time should be able to accept new observations as they arrive and not forget what they have learned~\cite{P7}. These features render financial forecasting quite different from traditional stationary sequence prediction and necessitate models that can simultaneously model nonlinear temporal dynamics, adjust in response to evolving market regimes and express uncertainty in future forecasts.

Classical recurrent architectures such as recurrent neural networks (RNNs), long short-term memory (LSTM) networks and gated recurrent units (GRUs) have been widely used for financial sequence modelling and are able to capture temporal dependencies from past observations~\cite{P20}. In more recent years, quantum machine learning (QML) has been proposed as an alternative type of computation, in which parameterized quantum circuits (PQCs) use quantum state representations, superposition and entanglement to model complex nonlinear relationships~\cite{P17, gandhudi2024explainable}. Hybrid quantum-classical forecasting models generally go hand in hand with variational quantum circuits, which act on a compact latent representation extracted classically~\cite{P1}. Although recent advances have been made, the current strategies are based on a fixed market distribution with a static quantum circuit configuration\cite{ahmad2026quantum}. As a result, they fail to explicitly adjust circuit complexity to varying volatility regimes, and must be conventionally retrained with new observations as they arrive. In sequential retraining, the parameters from previous market dynamics can be overwritten, resulting in catastrophic forgetting~\cite{P18, iyer2024artificial}. In addition, most hybrid QML forecasting methods involve point forecasts and fail to offer calibrated forecasts of predictive uncertainty, thus restricting their use during volatile and rapidly changing markets~\cite{P19}.

This paper introduces HQNN-CL, a hybrid quantum/classical continual learning framework for non-stationary forecasting in the financial market. The proposed architecture is based on first identifying the latent market states with a Gaussian Hidden Markov Model (HMM) applied to log-return and realized-volatility observations. A bidirectional LSTM is then used to encode the historical sequences in a quantum compatible small feature space with a regime conditioned bidirectional LSTM. This representation is then fed into an adaptive-depth parameterized quantum circuit that has different active entanglement depths, depending on the detected volatility regime, so that the circuit can transform shallower during stable periods and have a greater representational power during complex high-volatility conditions. For sequential market adaptation, Elastic Weight Consolidation (EWC) and experience replay are combined on the classical network parameters as well as on the quantum rotation angles in order to maintain the dynamics of previous time steps and introduce new observations. A probabilistic prediction head is finally designed to model the aleatoric and epistemic uncertainty, using heteroscedastic variance estimation and Monte Carlo Dropout, and then calibrating the resulting prediction intervals with a temperature. The prediction problem is posed on next-step log-returns and price levels are then reconstructed from the most recent observed closing price, thus avoiding a strongly trending, out-of-distribution price series for extrapolation. The main contributions of this work are summarized as follows:

\begin{itemize}
    \item To solve the non-stationary financial forecasting problem, this work suggests a regime-adaptive hybrid quantum-classical architecture: HQNN-CL, which combines market regime detection, temporal sequence encoding and quantum feature transformation.

    \item First, this work proposes to use a regime-adaptive parameterized quantum circuit that adaptively changes its active depth based on the market volatility that has been detected, allowing for controllable quantum representational capacity under different market regimes.

    \item Then, introduces experience replay to a shared classical-quantum parameter space, allowing for sequential model updates with restricted catastrophic forgetting between various, temporally evolving financial tasks.

    \item Finally, formulates a calibrated probabilistic forecasting mechanism that accounts for both the aleatoric and the epistemic uncertainty, and that gives point forecasts but also produces empirically calibrated prediction intervals.

\end{itemize}

The rest of this paper is structured as follows. The related literature on financial forecasting and continual learning and hybrid quantum computing is discussed in Section ~\ref{LR}. The proposed HQNN-CL framework and methodology are presented in Section ~\ref{PM}. The experiments setup, datasets, baseline models, and evaluation metrics are described in section ~\ref{ES}. Experimental results for ablation, uncertainty, and continual learning are reported and discussed in Section ~\ref{RD}. Finally in Section 6, the paper is concluded and future research directions are outlined.

\section{Literature Review}
\label{LR}
Forecasting a time series of financial data has received much attention, because forecasting plays an important role in making investment decisions and managing risk. While current methods range from recurrent neural networks to Transformer-based models and recent hybrid quantum-classical methods, most of them assume a stationary learning setting, and fail to account for non-stationary market dynamics, regime shifts, and catastrophic forgetting. This section briefly summarizes the recent development of financial forecasting, continual learning, and hybrid quantum computing and outlines the research gap the proposed HQNN-CL framework aims to solve.

\subsection{Financial Time-Series Forecasting}
Time-series forecasting is an area that has received a lot of research attention, due to its importance in portfolio management, algorithmic trading, and financial risk analysis. While classical statistical models such as ARIMA, GARCH and Hidden Markov Models (HMMs) have proven to be successful in representing linear temporal dynamics and volatility clustering, they are constrained by the assumption that they have to model stationary behavior, which is at odds with the high nonlinearity and rapid evolution of modern financial markets\cite{iyer2024artificial}. For this reason, machine learning approaches such as Support Vector Machines, Random Forests, and gradient boosting algorithms have proven to be a popular alternative because they can learn nonlinear relationships directly from past observations~\cite{P3, arora2025hybrid}. However, these methods are mostly based on hand-crafted features and have restricted ability to capture long-range temporal dependencies in sequential financial data. Financial forecasting has significantly improved with the use of deep learning architectures, which have the ability to learn automatically hierarchical temporal representations from historical observations of the market. In the realm of financial data for prediction, Recurrent Neural Networks (RNNs), Long Short-Term Memory (LSTM) networks, Bidirectional LSTMs, and more recently Transformer-based architectures have shown better performance~\cite{P9, deshpande2025prediction}. However, most deep learning methods rely on a stationary training distribution, and use the same model structure throughout training. As a result, it is often the case that the parameters of a model that are optimized for one market condition may deteriorate when market dynamics change, leading to poor generalization under a distribution shift and unstable market environments~\cite{P3} \cite{P11}.

\subsection{Hybrid Quantum-Classical Financial Forecasting}
Hybrid quantum-classical architectures which leverage the use of parameterized quantum circuits (PQCs) to provide improved nonlinear feature representation while preserving the optimization capability of classical neural networks have recently been developed in the field of quantum machine learning (QML)~\cite{P3, arora2025hybrid}. Previous research has combined variational quantum circuits with LSTM, recurrent neural networks, deep quantum networks, as well as reinforcement learning frameworks for stock price prediction and financial decision-making\cite{deshpande2025prediction, weinberg2025hybrid}. Representative examples of such networks are QLSTM, QuantumLeap, HQNN-FSP, frequency-domain recurrent quantum networks, QADQN, and contextual quantum neural networks, all showing the promise of quantum representations for financial forecasting~\cite{P6} \cite{P11}. While these models always surpass competing classical models, most of these use the same variational quantum circuit for each market observation regardless of market conditions~\cite{P18, singh2026stock, weinberg2025hybrid}. All three aspects of the quantum circuit, namely circuit depth, entanglement structure and encoding of features are fixed during the training and inference process, which would imply that the same quantum computation power is equally suitable for stable and highly volatile market periods~\cite{P1}\cite{P2}. Moreover, most of the current hybrid configurations are focused on improving the point prediction accuracy, and often lack the ability to continuously adapt or quantify uncertainty or even preserve sequential knowledge after deployment. Thus, existing quantum financial forecasting systems have limited real-world use with the always changing financial landscape~\cite{P11}\cite{P4}.

\subsection{Market Regime Modeling in Financial Prediction}
Financial markets naturally switch between different volatility regimes that have significantly different statistical properties. One of the most popular probabilistic methods for detecting hidden market conditions is the Hidden Markov Model (HMM), which, in addition to keeping the temporal dependence, explicitly captures the transitions between hidden regimes~\cite{P14}. By distinguishing the stable, transition and turbulent market behaviors, HMM-based forecasting has shown improvement in the estimation of the volatility, portfolio allocation and prediction of market trend~\cite{P8, singh2026stock}. This concept has been further developed more recently by Hidden Quantum Markov Models (HQMMs) which represent the states of a Markov model using quantum states and which have shown better predictive power than traditional HMMs for various financial indexes. Despite these developments, current regime-aware methods mainly use the regime detected as an extra feature to the learning forecast and not as an inherent part of the learning architecture~\cite{P5, ahmad2026quantum}. In particular, existing hybrid quantum neural networks (HQNNs) do not "self-regulate" their quantum computational complexity depending on the detected market state. As a result, the same quantum circuit settings are run for both low volatility and high turbulence, which is not efficient in the allocation of quantum representational capacity to various financial environments~\cite{P4}.

\subsection{Continual Learning for Sequential Financial Markets}
Financial forecasting models are subject to continuously changing environments where new observations are received sequentially and historical data distributions change over time\cite{P3}~\cite{P10}. A conventional supervised learning approach would require all the training data to be available prior to deployment, and then would need to be retrained with new observations added to the dataset~\cite{P12}. This retraining is very costly in terms of computation and often leads to catastrophic forgetting (where sequential optimization alters parameters that are significant for previous market conditions)~\cite{P15}. A promising solution to tackle the sequential adaptation is the continual learning, which aims at maintaining previously learned knowledge while acquiring new knowledge\cite{P13}~\cite{P8}. Regularization based methods like Elastic Weight Consolidation (EWC), replay-based learning, parameter isolation and memory rehearsal have been very successful in computer vision and natural language processing. But in the context of hybrid quantum-classical financial forecasting, their use has been hardly explored~\cite{P1}. The current quantum forecasting models are trained offline with static data, and there is no method to keep the quantum representation after the sequential evolution of the market. Therefore, the real-life implementation of quantum financial models in the financial landscape under the current variable market conditions is an ongoing research challenge~\cite{P14}.

\subsection{Probabilistic Forecasting and Research Gap}
New studies on financial forecasting have more and more focused on the use of a 'probabilistic' forecast, as opposed to deterministic point forecast, which enable investors to express predictive uncertainty when taking decisions. In financial forecasting, there are several methods proposed to estimate two sources of uncertainty, namely epistemic uncertainty and aleatoric uncertainty, such as Bayesian neural networks, Monte Carlo Dropout, and heteroscedastic regression~\cite{P15}. Most existing quantum machine learning models, however, only minimize deterministic regression losses and rarely output calibrated confidence intervals with predicted prices. Furthermore, uncertainty estimation has not yet been incorporated in an adaptive quantum architecture or in a framework of continual learning\cite{singh2026stock}.

From the above analysis, there are four important research gaps. First, current hybrid quantum-classical forecasting models are based on fixed-depth parameterized quantum circuits, which do not consider how to adjust computational complexity in response to the varying market volatility. Second, detected market regimes seldom are used in both classical temporal encoding and quantum feature transformation. Third, learning mechanisms that can maintain the classical neural parameters and the quantum circuit parameters in sequential learning tasks in the financial field have not been studied. Lastly, existing quantum financial forecasting models tend to give a determinate prediction with no calibrated uncertainty estimation. In an attempt to overcome these limitations, this work proposes a regime-adaptive hybrid quantum-classical continual learning framework, called HQNN-CL, which consists of Gaussian HMM-based market regime detection, regime-conditioned BiLSTM encoding, adaptive-depth parameterized quantum circuits, joint classical-quantum continual learning using elastic weight consolidation and experience replay, and calibrated probabilistic forecasting for robust sequential financial prediction.

\section{Proposed Methodology}
\label{PM}

\begin{figure}[!htbp]
 \centering
 \makebox[\textwidth][c]{\includegraphics[width=1\textwidth]{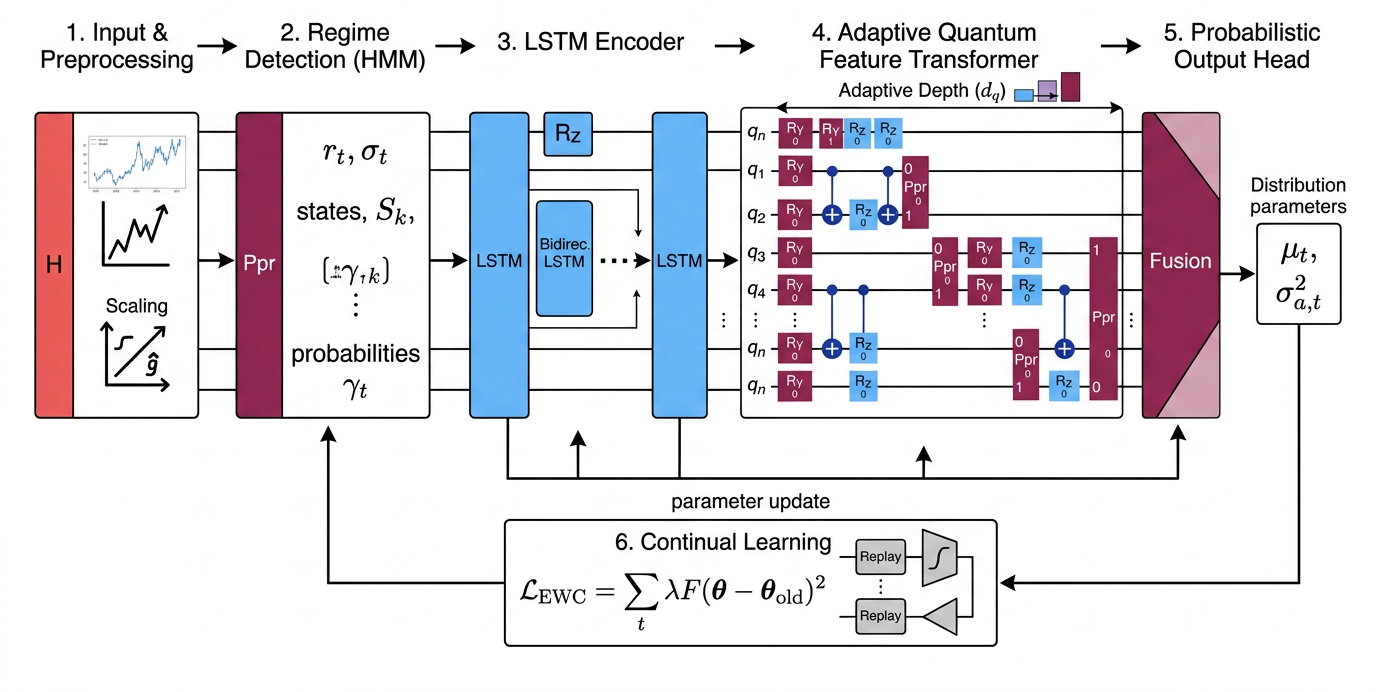}}%
 \caption{HQNN-CL architecture. The framework integrates HMM-based regime detection, BiLSTM temporal encoding, regime-adaptive quantum feature transformation, probabilistic forecasting, and continual learning with EWC and experience replay.}
 \label{fig-1}
\end{figure}

The proposed HQNN-CL framework integrates regime-aware temporal modeling, adaptive quantum feature transformation, continual learning, and calibrated probabilistic forecasting for non-stationary financial time series. The complete architecture is illustrated in Figure~\ref{fig-1}. Historical market observations are first transformed into stationary return-based sequences and augmented with technical indicators. A Gaussian Hidden Markov Model (HMM) identifies latent volatility regimes from return and realized-volatility dynamics. These regime states condition a bidirectional long short-term memory (BiLSTM) network that extracts temporal representations from historical observations. The resulting latent features are encoded into a parameterized quantum circuit (PQC), whose active depth changes with the detected market regime. Pauli-$Z$ expectation measurements produce quantum features that are fused with the classical representation for probabilistic return forecasting. Finally, Elastic Weight Consolidation (EWC) and experience replay enable sequential model updates while limiting catastrophic forgetting across temporally evolving financial tasks.

\subsection{Financial time-series formulation}

For a financial asset, the market observation at trading time $t$ is represented by the open price $O_t$, high price $H_t$, low price $L_t$, closing price $C_t$, and trading volume $V_t$. The corresponding observation vector is defined as ~\ref{eq:input}

\begin{equation}
\mathbf{x}_t =
[O_t,H_t,L_t,C_t,V_t]^{\top}.
\label{eq:input}
\end{equation}

Given a historical lookback window of length $T$, the input sequence is
$\mathbf{X}^{(T)}_t=
[\mathbf{x}_{t-T},\ldots,\mathbf{x}_{t-1}]$.
The objective is to learn a parameterized forecasting model
$f_{\Theta}(\mathbf{X}^{(T)}_t)$, where $\Theta$ represents the complete set of classical and quantum trainable parameters.

Direct forecasting of raw closing prices becomes problematic for strongly trending equities because chronological test observations may lie outside the price interval observed during training. A MinMax-scaled model can therefore be forced to extrapolate beyond its learned output range. To avoid this behavior, HQNN-CL predicts the next-step logarithmic return as ~\ref{eq:return}

\begin{equation}
r_t =
\ln\left(
\frac{C_{t+1}}{C_t}
\right).
\label{eq:return}
\end{equation}

Log-returns represent relative price movements and exhibit substantially less dependence on the absolute price level. Let $\hat{\sigma}_r$ denote the standard deviation of returns estimated exclusively from the training data. The forecasting target is scaled as
$\tilde{r}_t=r_t/(5\hat{\sigma}_r)$, which approximately maps ordinary market movements to the interval $[-1,1]$. After prediction, the closing price is reconstructed from the most recently observed price as ~\ref{eq:price_reconstruction}

\begin{equation}
\hat{C}_{t+1}
=
C_t
\exp\left(
5\hat{\sigma}_r
\hat{\tilde{r}}_t
\right).
\label{eq:price_reconstruction}
\end{equation}

This formulation anchors every prediction to $C_t$ and avoids requiring the model to extrapolate absolute price levels. Consequently, the learning problem is performed in stationary return space, whereas final forecasting performance remains interpretable in original price units.

\subsection{Financial feature construction}

The OHLCV observations are augmented with momentum, trend, and volatility information. In particular, HQNN-CL incorporates the Relative Strength Index (RSI), Moving Average Convergence Divergence (MACD), Average Directional Index (ADX), 5-day and 20-day simple moving averages, and realized volatility. Realized volatility is computed from the most recent $w=10$ log-returns as ~\ref{eq:volatility}

\begin{equation}
\sigma_t =
\operatorname{Std}
(r_{t-w+1},\ldots,r_t).
\label{eq:volatility}
\end{equation}

RSI is calculated over a 14-day period using the average positive and negative price changes, whereas MACD is obtained from the difference between the 12-day and 26-day exponential moving averages. ADX is used to represent directional trend strength independently of trend direction. These indicators provide complementary information describing short-term momentum, persistent price movement, and local market variability.

The complete candidate feature set contains the raw market variables and derived technical indicators. To reduce redundant inputs before temporal and quantum processing, univariate $F$-regression is applied against the log-return target. The five highest-ranked variables are retained and normalized to $[-1,1]$ using parameters estimated only from the training partition. This dimensionality is consistent with the compact latent representation required by the subsequent quantum stage. A sliding window of $T=10$ trading days is used, yielding
$\mathbf{X}_t\in\mathbb{R}^{10\times5}$.
Training, validation, and testing observations are partitioned chronologically without random shuffling across temporal boundaries.

\subsection{Latent market regime detection}

Financial time series frequently exhibit transitions between periods of stable movement and intervals of elevated volatility. A model that treats all observations as realizations from a single stationary distribution may fail to distinguish these structurally different conditions. HQNN-CL therefore introduces an explicit Gaussian HMM to infer latent market states.

At time $t$, the HMM receives the bivariate observation
$\mathbf{o}_t=[r_t,\sigma_t]^{\top}$.
Let $z_t\in\{0,1,2\}$ denote the hidden market regime. The transition probability between states is defined as ~\ref{eq:hmm_transition}

\begin{equation}
A_{ij}
=
P(z_t=j\mid z_{t-1}=i),
\label{eq:hmm_transition}
\end{equation}

and each latent state $k$ generates observations according to the full-covariance Gaussian emission model ~\ref{eq:hmm_emission}

\begin{equation}
p(\mathbf{o}_t\mid z_t=k)
=
\mathcal{N}
\left(
\mathbf{o}_t;
\boldsymbol{\mu}_k,
\boldsymbol{\Sigma}_k
\right).
\label{eq:hmm_emission}
\end{equation}

The HMM parameters
$\{\boldsymbol{\pi},\mathbf{A},
\boldsymbol{\mu}_k,\boldsymbol{\Sigma}_k\}$
are estimated using the Baum--Welch expectation-maximization algorithm. Viterbi decoding subsequently identifies the most likely state sequence. Because HMM state indices have no intrinsic ordering, the states are sorted according to their estimated mean volatility,
$\mu_0^{(\sigma)}<
\mu_1^{(\sigma)}<
\mu_2^{(\sigma)}$.
State $0$ therefore denotes the low-volatility regime, state $1$ denotes a transition regime, and state $2$ represents high volatility.

The inferred state is used structurally rather than merely being appended as another financial feature. It conditions the classical sequence representation and determines the number of active quantum variational layers. This creates a direct connection between the detected statistical condition of the market and the computational capacity allocated by the forecasting architecture.

\subsection{Regime-conditioned BiLSTM encoder}

HQNN-CL uses a shared regime-conditioned BiLSTM to extract temporal dependencies. Independent recurrent networks for individual market states may reduce sample efficiency because transition and high-volatility regimes generally contain fewer observations. A shared encoder instead allows information to propagate across all market states while a learned embedding preserves regime-specific context.

Let
$e(z_t)\in\mathbb{R}^{d_e}$
represent the trainable embedding associated with regime $z_t$, where $d_e=4$. The regime embedding is concatenated with the financial feature vector at every timestep~\ref{eq:regime_embedding},

\begin{equation}
\tilde{\mathbf{x}}_j =
[\mathbf{x}_j;e(z_t)].
\label{eq:regime_embedding}
\end{equation}

The augmented sequence is processed in forward and reverse temporal directions. The resulting representation is written as ~\ref{eq:bilstm}

\begin{equation}
\mathbf{h}_j =
\left[
\operatorname{LSTM}_{f}
(\tilde{\mathbf{x}}_j,
\overrightarrow{\mathbf{h}}_{j-1});
\operatorname{LSTM}_{b}
(\tilde{\mathbf{x}}_j,
\overleftarrow{\mathbf{h}}_{j+1})
\right].
\label{eq:bilstm}
\end{equation}

The encoder contains two BiLSTM layers with 32 hidden units in each direction and an inter-layer dropout rate of $0.15$. The final bidirectional state
$\mathbf{h}_T\in\mathbb{R}^{64}$
summarizes the temporal context over the ten-day input window.

Before quantum processing, the recurrent representation is compressed into a four-dimensional bounded latent vector~\ref{eq:projection}

\begin{equation}
\mathbf{z}
=
\tanh
\left(
\mathbf{W}_p\mathbf{h}_T+\mathbf{b}_p
\right),
\qquad
\mathbf{z}\in[-1,1]^4.
\label{eq:projection}
\end{equation}

The hyperbolic tangent activation is used because the resulting bounded values can be directly transformed into valid quantum rotation angles. This projection also creates a compact information bottleneck between the classical temporal encoder and the four-qubit PQC.

\subsection{Adaptive quantum feature transformation}

The quantum component receives the latent vector $\mathbf{z}$ produced by the BiLSTM encoder. Similar to the quantum architecture adopted in HQNN-FSP, the circuit is initialized in the computational ground state and classical information is incorporated through angle encoding. :contentReference[oaicite:2]{index=2} However, the proposed HQNN-CL modifies the quantum transformation by introducing regime-dependent circuit depth.

\subsubsection{Quantum state preparation}

The four-qubit quantum register is initialized as~\ref{eq:ground_state}

\begin{equation}
|\psi^{(0)}\rangle
=
|0\rangle^{\otimes N_q},
\qquad N_q=4.
\label{eq:ground_state}
\end{equation}

For latent component $z_i$, the rotation angles are defined as
$\theta_i=\arcsin(\operatorname{clip}(z_i,-0.999,0.999))$
and
$\phi_i=\arccos(\operatorname{clip}(z_i,-0.999,0.999))$.
The corresponding single-qubit encoding transformation is~\ref{eq:rotation}

\begin{equation}
R_Y(\theta_i)
=
e^{-i\theta_iY/2},
\qquad
R_Z(\phi_i)
=
e^{-i\phi_iZ/2}.
\label{eq:rotation}
\end{equation}

Here, $Y$ and $Z$ denote the Pauli-$Y$ and Pauli-$Z$ matrices, respectively. The $R_Y$ transformation modifies the probability amplitudes of the computational basis states, whereas $R_Z$ introduces phase information. Applying both rotations allows each classical feature to influence both amplitude and phase components of the quantum state.

The encoded state is therefore obtained by~\ref{eq:encoding}

\begin{equation}
|\psi(\mathbf{z})\rangle
=
U_{\mathrm{enc}}(\mathbf{z})
|0\rangle^{\otimes N_q},
\label{eq:encoding}
\end{equation}

where
$U_{\mathrm{enc}}(\mathbf{z})
=
\bigotimes_{i=1}^{N_q}
R_Z(\phi_i)R_Y(\theta_i)$.
All angle calculations are implemented as differentiable tensor operations so that gradients from the forecasting objective propagate through the quantum state preparation stage and update the preceding BiLSTM encoder.

\subsubsection{Customized variational ansatz}

After angle encoding, the quantum state is transformed through a layered variational ansatz. Each layer contains trainable $R_Y$ and $R_Z$ rotations followed by a cyclic CNOT entanglement pattern. For layer $d$, the parameterized transformation is represented by~\ref{eq:quantum_layer}

\begin{equation}
U_d(\mathbf{w}_d)
=
U_{\mathrm{ent}}
\left[
\bigotimes_{i=1}^{N_q}
R_Z(w_{d,i,2})
R_Y(w_{d,i,1})
\right],
\label{eq:quantum_layer}
\end{equation}

where $\mathbf{w}_d$ contains the trainable rotation angles of layer $d$. The entanglement operator applies
$\operatorname{CNOT}_{0,1}$,
$\operatorname{CNOT}_{1,2}$,
$\operatorname{CNOT}_{2,3}$, and
$\operatorname{CNOT}_{3,0}$,
forming a ring structure between the four qubits.

The corresponding effective generator follows the Hamiltonian-inspired formulation used in customized financial QNNs and can be expressed as~\ref{eq:hamiltonian}

\begin{equation}
H =
\sum_i
\left(
\frac{\theta_i}{2}\sigma_y^{(i)}
+
\frac{\phi_i}{2}\sigma_z^{(i)}
\right)
+
\sum_{\langle i,j\rangle}
\frac{1-\sigma_z^{(i)}}{2}
\sigma_x^{(j)}.
\label{eq:hamiltonian}
\end{equation}

The first term describes local feature-dependent rotations, whereas the second represents controlled interactions between neighboring qubits. In contrast to an unentangled quantum feature map, the ring-connected ansatz allows information encoded on one qubit to influence measurements performed on other qubits. The resulting quantum representation therefore models interactions between components of the temporal latent vector.

\subsubsection{Regime-adaptive circuit depth}

Conventional PQCs generally execute a fixed number of variational layers for every input. HQNN-CL instead assigns quantum depth according to the detected market regime. The active depth is defined as ~\ref{eq:adaptive_depth}

\begin{equation}
D(z_t)
=
\min
\left(
2+z_t,D_{\max}
\right),
\qquad
D_{\max}=5.
\label{eq:adaptive_depth}
\end{equation}

Consequently, low-volatility observations use two quantum layers, transition observations use three layers, and high-volatility observations activate four layers. The transformed state is ~\ref{eq:quantum_state}

\begin{equation}
|\psi^{(D)}\rangle
=
\prod_{d=1}^{D(z_t)}
U_d(\mathbf{w}_d)
|\psi(\mathbf{z})\rangle.
\label{eq:quantum_state}
\end{equation}

The objective of adaptive depth is not to assume that deeper circuits are universally superior. Instead, the model allocates additional parameterized rotations and entanglement operations when the HMM identifies a statistically more volatile market state. Stable observations are processed through a shallower circuit, reducing unnecessary transformation depth, whereas high-volatility states receive greater quantum representational capacity.

All $D_{\max}$ parameter blocks belong to a single shared PQC. The deeper layers are updated when transition or high-volatility observations activate them. This shared formulation is important for continual learning because EWC can estimate the importance of individual quantum parameters and constrain only those rotations that are relevant to previously observed financial tasks.

\subsubsection{Quantum measurement}

Following the variational transformation, every qubit is measured using the Pauli-$Z$ observable. The quantum feature associated with qubit $i$ is ~\ref{eq:measurement}

\begin{equation}
q_i
=
\langle
\psi^{(D)}
|
Z_i
|
\psi^{(D)}
\rangle.
\label{eq:measurement}
\end{equation}

The four expectation values form the quantum feature vector
$\mathbf{q}=[q_1,\ldots,q_4]^{\top}\in[-1,1]^4$.
Unlike measurement-frequency reconstruction, expectation values provide continuous differentiable features that can be directly supplied to the subsequent classical layer. The quantum circuit is implemented using PennyLane's state-vector simulator and PyTorch automatic differentiation. Batched quantum broadcasting is used to execute multiple encoded sequences without a sample-wise Python circuit loop.

\subsection{Classical-quantum fusion and probabilistic output}

The classical latent vector $\mathbf{z}$ retains the temporal representation generated by the BiLSTM, whereas $\mathbf{q}$ represents the same information after nonlinear quantum transformation and entanglement. HQNN-CL preserves both representations rather than replacing the classical feature vector with quantum measurements.

The representations are concatenated and mapped into a 16-dimensional fusion space ~\ref{eq:fusion},

\begin{equation}
\mathbf{f}
=
\operatorname{ReLU}
\left(
\operatorname{Dropout}
\left(
\mathbf{W}_f
[\mathbf{z};\mathbf{q}]
+\mathbf{b}_f
\right)
\right).
\label{eq:fusion}
\end{equation}

The dropout probability is set to $0.15$. The fused vector is subsequently processed by two independent linear heads. One head estimates the conditional mean $\mu_t$, whereas the second predicts the log-variance $\log\sigma_t^2$. Hence~\ref{eq:dropout},

\begin{equation}
    p(\tilde{r}_t\mid\mathbf{X}_t)
=
\mathcal{N}
(\mu_t,\sigma_t^2).
\label{eq:dropout}
\end{equation}

The network is optimized using the heteroscedastic Gaussian negative log-likelihood~\ref{eq:nll}

\begin{equation}
\mathcal{L}_{\mathrm{NLL}}
=
\frac{1}{2}\log\sigma_t^2
+
\frac{
(\tilde{r}_t-\mu_t)^2
}{
2\sigma_t^2
}.
\label{eq:nll}
\end{equation}

Unlike deterministic mean squared error, the NLL objective jointly learns the forecast and its input-dependent uncertainty. Large prediction errors are associated with increased estimated variance, whereas the logarithmic penalty prevents the network from arbitrarily increasing $\sigma_t^2$. The learned variance therefore represents the aleatoric component of uncertainty associated with intrinsic financial noise.

\subsection{Continual hybrid quantum-classical learning}

Financial models trained once on a historical interval gradually become outdated as market dynamics evolve. To simulate periodic model updating, the chronological training data are divided into three sequential tasks,
$\mathcal{T}_1\prec\mathcal{T}_2\prec\mathcal{T}_3$.
HQNN-CL is trained on $\mathcal{T}_1$ before receiving $\mathcal{T}_2$ and $\mathcal{T}_3$ in temporal order.

Let
$\Theta=\{\Theta_C,\Theta_Q\}$,
where $\Theta_C$ denotes all classical parameters and $\Theta_Q$ contains the quantum rotation angles $\mathbf{w}$. Sequential optimization of $\Theta$ can overwrite parameter configurations important to earlier periods. HQNN-CL combines EWC and experience replay to reduce this effect.

\subsubsection{Elastic Weight Consolidation}

After learning task $\mathcal{T}_k$, the empirical Fisher information estimates the sensitivity of the forecasting loss to parameter $\Theta_i$ ~\ref{eq:fisher},

\begin{equation}
F_i^{(k)}
=
\mathbb{E}_{(\mathbf{X},y)\sim\mathcal{T}_k}
\left[
\left(
\frac{
\partial\mathcal{L}_{\mathrm{NLL}}
}{
\partial\Theta_i
}
\right)^2
\right].
\label{eq:fisher}
\end{equation}

A large $F_i^{(k)}$ indicates that small changes in parameter $\Theta_i$ strongly influence the forecasting objective for task $\mathcal{T}_k$. The optimized parameter values
$\Theta_i^{*(k)}$
and their importance estimates are stored after each task.

During subsequent learning, parameter displacement is penalized using ~\ref{eq:ewc}

\begin{equation}
\mathcal{L}_{\mathrm{EWC}}
=
\lambda
\sum_{k<t}
\sum_i
F_i^{(k)}
\left(
\Theta_i-\Theta_i^{*(k)}
\right)^2,
\label{eq:ewc}
\end{equation}

where $\lambda=500$ controls consolidation strength. The penalty is applied to the complete hybrid parameter space. Thus, BiLSTM weights, projection parameters, fusion layers, and quantum rotation angles are protected according to their previous forecasting importance.

\subsubsection{Experience replay and joint objective}

EWC preserves important parameter directions but does not explicitly expose the network to previous market observations. Therefore, HQNN-CL additionally maintains a reservoir replay memory
$\mathcal{M}$ with a maximum capacity of 1,500 sequences. Representative observations from earlier tasks are retained through uniform reservoir replacement.

When learning a new task, a replay batch is sampled from $\mathcal{M}$ and optimized jointly with the current observations. The complete HQNN-CL training objective is ~\ref{eq:total_loss}

\begin{equation}
\mathcal{L}_{\mathrm{total}}
=
\mathcal{L}_{\mathrm{NLL}}^{\mathrm{new}}
+
\mathcal{L}_{\mathrm{EWC}}
+
\beta
\mathcal{L}_{\mathrm{NLL}}^{\mathrm{replay}},
\label{eq:total_loss}
\end{equation}

where $\beta=0.3$. EWC constrains destructive movement in important regions of the hybrid parameter space, whereas replay maintains direct exposure to previously observed financial sequences. The two mechanisms therefore provide complementary parameter-level and data-level forgetting mitigation.

All parameters are optimized jointly using AdamW with an initial learning rate of $1.5\times10^{-3}$ and weight decay of $10^{-4}$. A cosine-annealing learning-rate schedule is applied independently during each sequential task. The global gradient norm is clipped at $1.0$ to stabilize joint optimization across recurrent and quantum parameters.

\subsection{Calibrated predictive uncertainty}

The probabilistic output head directly estimates aleatoric uncertainty. However, uncertainty due to limited model knowledge must also be considered, particularly when the market enters conditions that differ from previously observed data. HQNN-CL estimates this epistemic component using Monte Carlo Dropout.

At inference time, dropout remains active and $S=50$ stochastic forward passes are performed. Let
$\{\mu_t^{(s)},\sigma_t^{2(s)}\}_{s=1}^{S}$
denote the resulting forecasts. The total predictive variance is obtained from ~\ref{eq:uncertainty}

\begin{equation}
\sigma_{\mathrm{total}}^2
=
\underbrace{
\operatorname{Var}_{s}
\left(
\mu_t^{(s)}
\right)
}_{\text{epistemic}}
+
\underbrace{
\frac{1}{S}
\sum_{s=1}^{S}
\sigma_t^{2(s)}
}_{\text{aleatoric}}.
\label{eq:uncertainty}
\end{equation}

The first term measures variation between stochastic model predictions and therefore approximates epistemic uncertainty. The second term averages the data-dependent variances predicted by the probabilistic output head.

Raw uncertainty estimates may not provide the desired empirical interval coverage. A scalar temperature $\tau^{*}$ is therefore selected on the validation set to minimize the difference between nominal and empirical $95\%$ coverage ~\ref{eq:calibration},

\begin{equation}
\tau^{*}
=
\arg\min_{\tau>0}
\left|
P
\left(
|y-\bar{\mu}|
\leq
1.96\tau\sigma_{\mathrm{total}}
\right)
-0.95
\right|.
\label{eq:calibration}
\end{equation}

The final prediction interval is
$\bar{\mu}_t\pm
1.96\tau^{*}\sigma_{\mathrm{total}}$.
After return-to-price reconstruction, uncertainty is propagated locally through the exponential transformation using its first-order derivative. HQNN-CL therefore generates both a next-step price estimate and a calibrated confidence interval.

The complete learning process operates sequentially. As new market data form the next temporal task, HQNN-CL updates the shared classical--quantum model using the replay-augmented EWC objective. The HMM continuously provides regime information to the temporal encoder and quantum-depth controller, while the calibrated probabilistic head quantifies uncertainty associated with each forecast. This integration allows the proposed architecture to adapt to evolving market dynamics without treating every market condition as statistically identical or requiring full retraining from initialization.

\section{Experimental Setup}
\label{ES}
This section presents the experimental setup for the proposed HQNN-CL framework. In order to ensure a complete and reproducible evaluation of the performance, it provides the complete list of datasets, data preprocessing steps, baseline models, implementation details, evaluation metrics, and experimental protocol.

\subsection{Datasets}
The suggested HQNN-CL framework was tested on four publicly available datasets of daily stock market data with different market characteristics such as large-cap technology stocks, region stocks, and range-bound stocks of financial instruments. The selected data sets include Microsoft (MSFT), Systems Limited (SYS), Teradata (TDC) and Tata Consultancy Services (TCS) with around ten years of historical trading data from July 2016 to May/June 2026. There are 2,443 to 2,486 trading records with daily Open, High, Low, Close, and Volume (OHLCV) observations in each set. The long-term trends of Microsoft and Tata Consultancy are upward and typical of well-established large-cap technology stocks, Systems Limited shows clear regime switches and regime dependent volatility, while Teradata is a relatively range-bound stock with lower trend persistence. Table~\ref{tab:datasets} provides the statistics of the datasets. To avoid data leakage from the future, chronological 70\%/10\%/20\% train/validation/test splits were used throughout all experiments. Input sequences were created using a fixed lookback window of 10 trading days, and all models were trained to predict the stationary next-step log-return target introduced in Section~2.1, first before reconstructing closing prices in the future for evaluation.

\subsection{Baseline Models and Training Configuration}
HQNN-CL was tested against 7 leading representative deep learning forecasting models, involving 3 recurrent models (LSTM, RNN, and GRU), and 4 state-of-the-art Transformer-based models (PatchTST, iTransformer, TimesNet, and FEDformer). For the recurrent baselines, use 64 hidden units in two hidden layers; and for PatchTST, set patch size to 5, $d_{\mathrm{model}}=32$ and number of Transformer encoder layers to 2, and number of attention heads to 4. Inverted Variate Attention is used in the iTransformer with $d_{\mathrm{model}}=32$; TimesNet leverages the top-3 temporal periods obtained by the FFT and the two-dimensional convolution blocks; and FEDformer applies the frequency-enhanced attention and the trend--seasonality decomposition techniques. All baselines were trained with the same stationary log-return formulation, chronological splits of the data, lookback window, AdamW optimizer with a learning rate of $3\times10^{-3}$, ReduceLROnPlateau, early stopping with a patience of 12, and a maximum of 60 training epochs. Thus, the difference in the forecasting ability can be attributed to the differences in the architecture design, but not to the differences in the preprocessing, optimization and training protocol.

\section{Results \& Discussion}
\label{RD}
In this section, show the experimental results of the proposed HQNN-CL framework and compare them with the state-of-the-art baseline models. From various angles such as forecasting accuracy, continual learning capacity, uncertainty estimation, regime adaptability and statistical significance the results are analyzed to show the effectiveness and robustness of the proposed approach.

\subsection{Gradient-Based Feature Importance}
\begin{figure}[!htbp]
\centering

\begin{subfigure}[t]{0.49\textwidth}
    \centering
    \includegraphics[width=\linewidth]{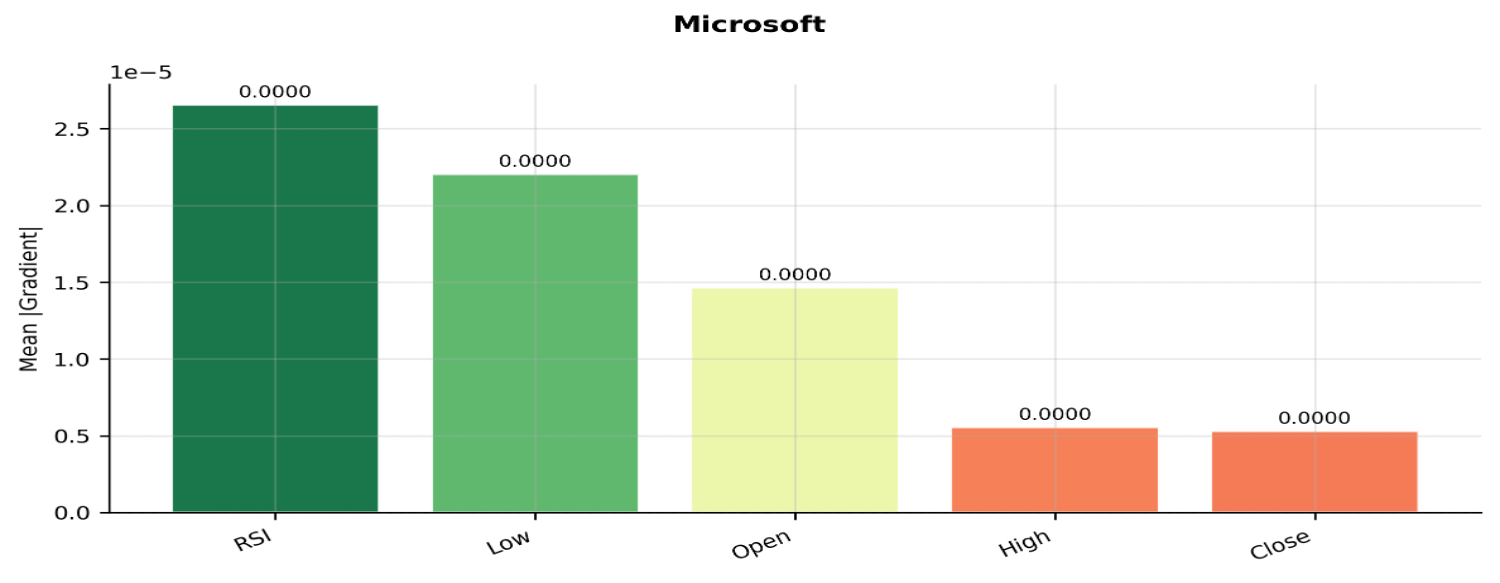}
    \caption{Microsoft (MSFT).}
    \label{fig:gb_msft}
\end{subfigure}
\begin{subfigure}[t]{0.49\textwidth}
    \centering
    \includegraphics[width=\linewidth]{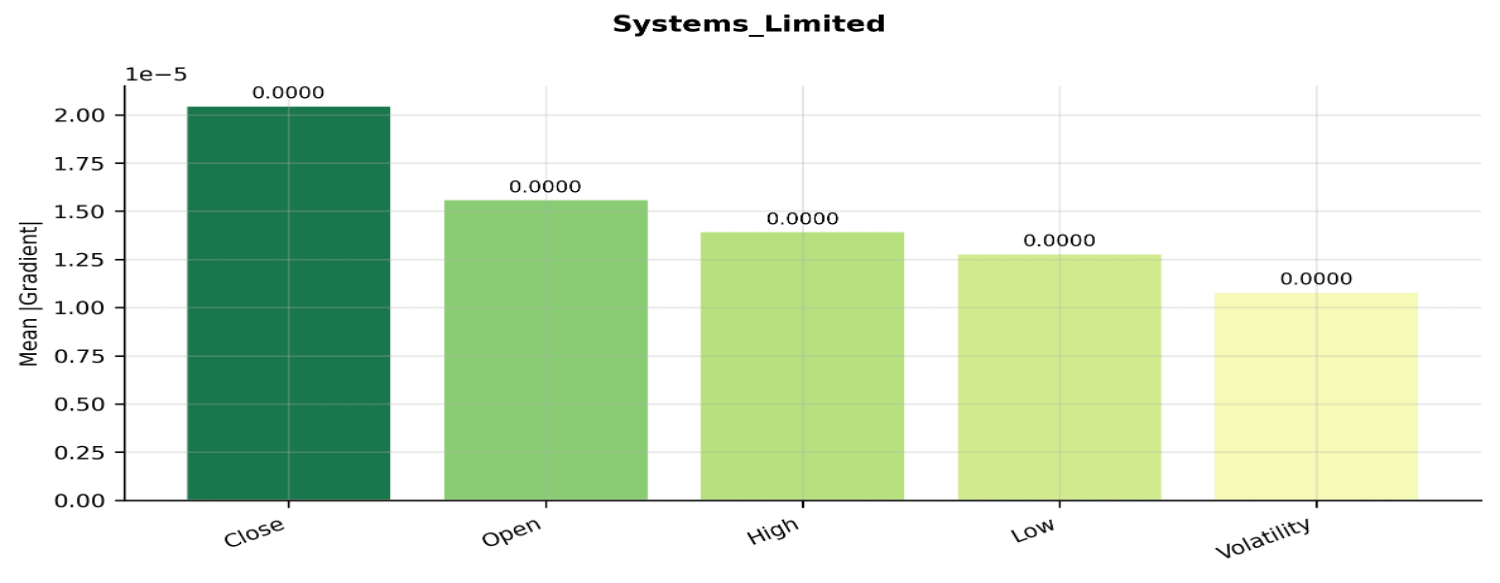}
    \caption{Systems Limited (SYS).}
    \label{fig:gb_sys}
\end{subfigure}

\vspace{0.35cm}

\begin{subfigure}[t]{0.49\textwidth}
    \centering
    \includegraphics[width=\linewidth]{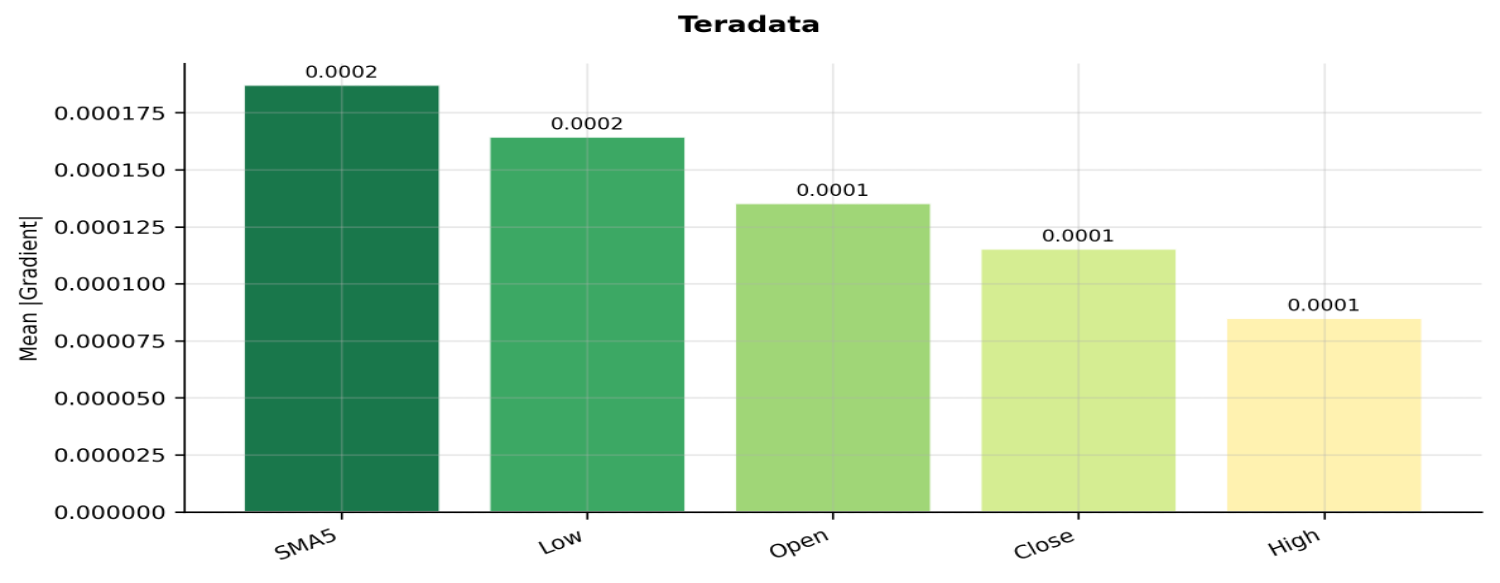}
    \caption{Teradata (TDC).}
    \label{fig:gb_tdc}
\end{subfigure}
\begin{subfigure}[t]{0.49\textwidth}
    \centering
    \includegraphics[width=\linewidth]{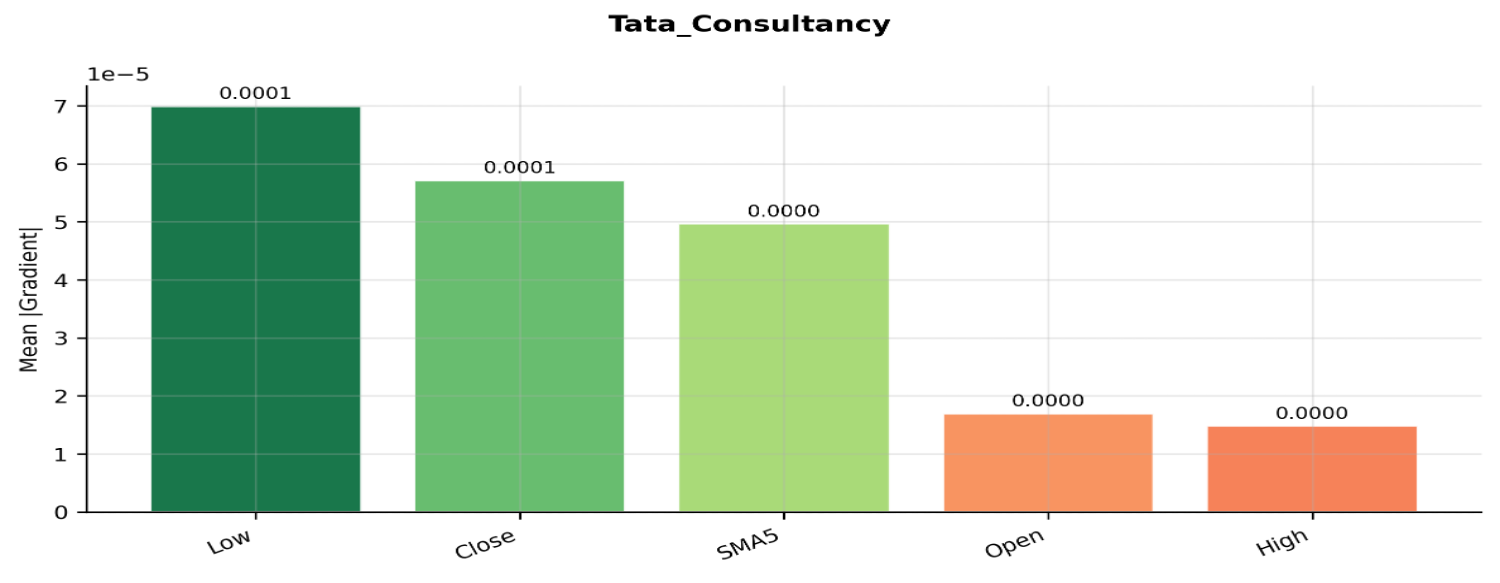}
    \caption{Tata Consultancy Services (TCS).}
    \label{fig:gb_tcs}
\end{subfigure}

\caption{\textbf{Gradient-based feature importance across the four benchmark datasets.} Mean absolute input gradients identify the most influential market variables, showing that HQNN-CL primarily relies on price-based features together with technical indicators such as RSI and SMA5 for forecasting.}
\label{fig:feature_importance}
\end{figure}

The mean absolute input gradients that were used to estimate the importance of the features for all four datasets are plotted in Figure~\ref{fig:feature_importance}. HQNN-CL always groups together the most important aspects of the chart, which are the price features (Open, Close, High, Low) and the most important technical indicators (RSI and SMA5). Microsoft is leading by RSI, Systems Limited by Close, Teradata by SMA5 and Tata Consultancy by Low prices. These findings reveal that HQNN-CL is able to extract interpretable, and economically meaningful representations and effectively utilise both market momentum and trend information for financial forecasting.

\subsection{Regime Analysis}
\begin{figure}[!htbp]
\centering

\begin{subfigure}[t]{0.49\textwidth}
    \centering
    \includegraphics[width=\linewidth]{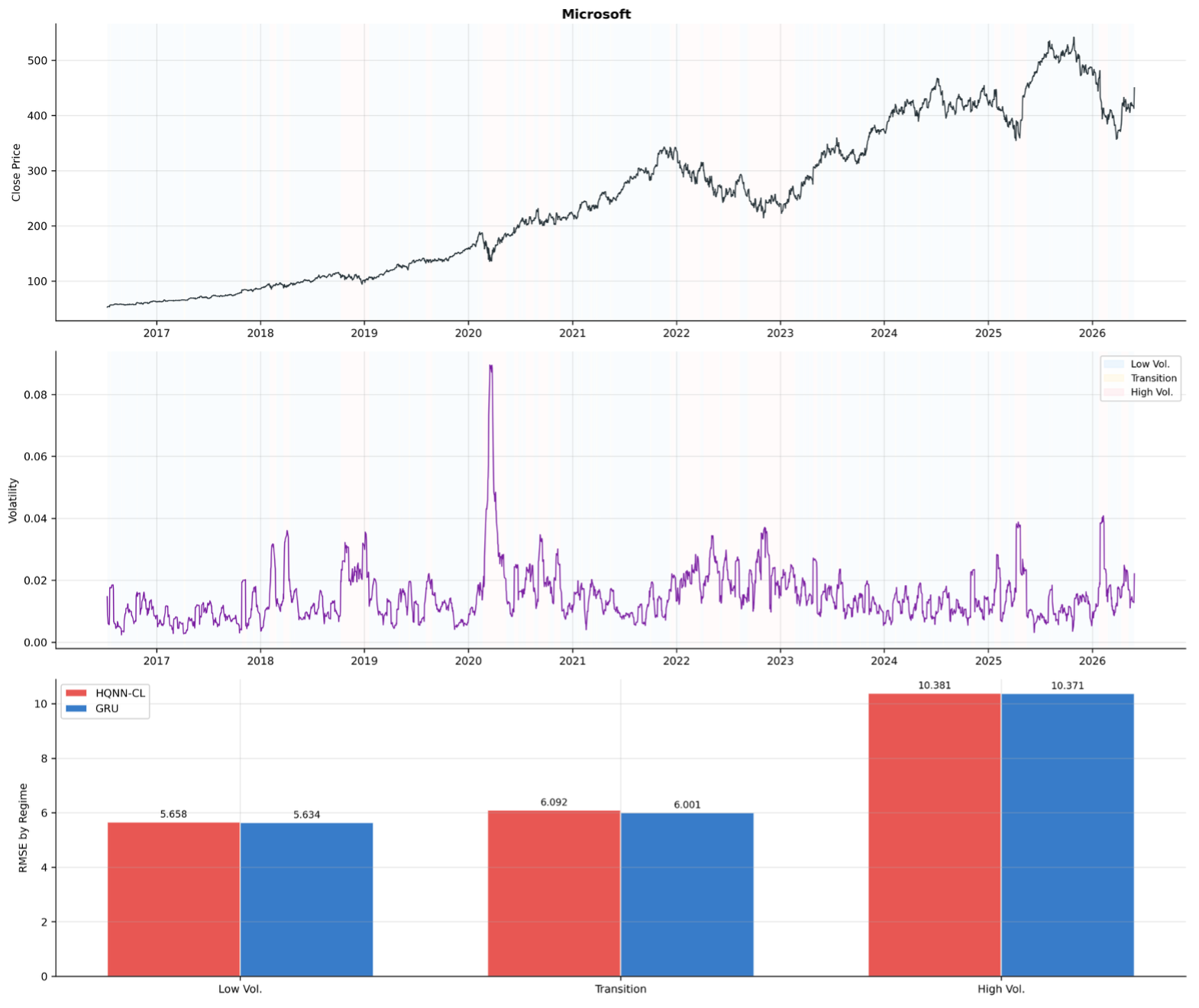}
    \caption{Microsoft (MSFT).}
    \label{fig:regime_msft}
\end{subfigure}
\begin{subfigure}[t]{0.49\textwidth}
    \centering
    \includegraphics[width=\linewidth]{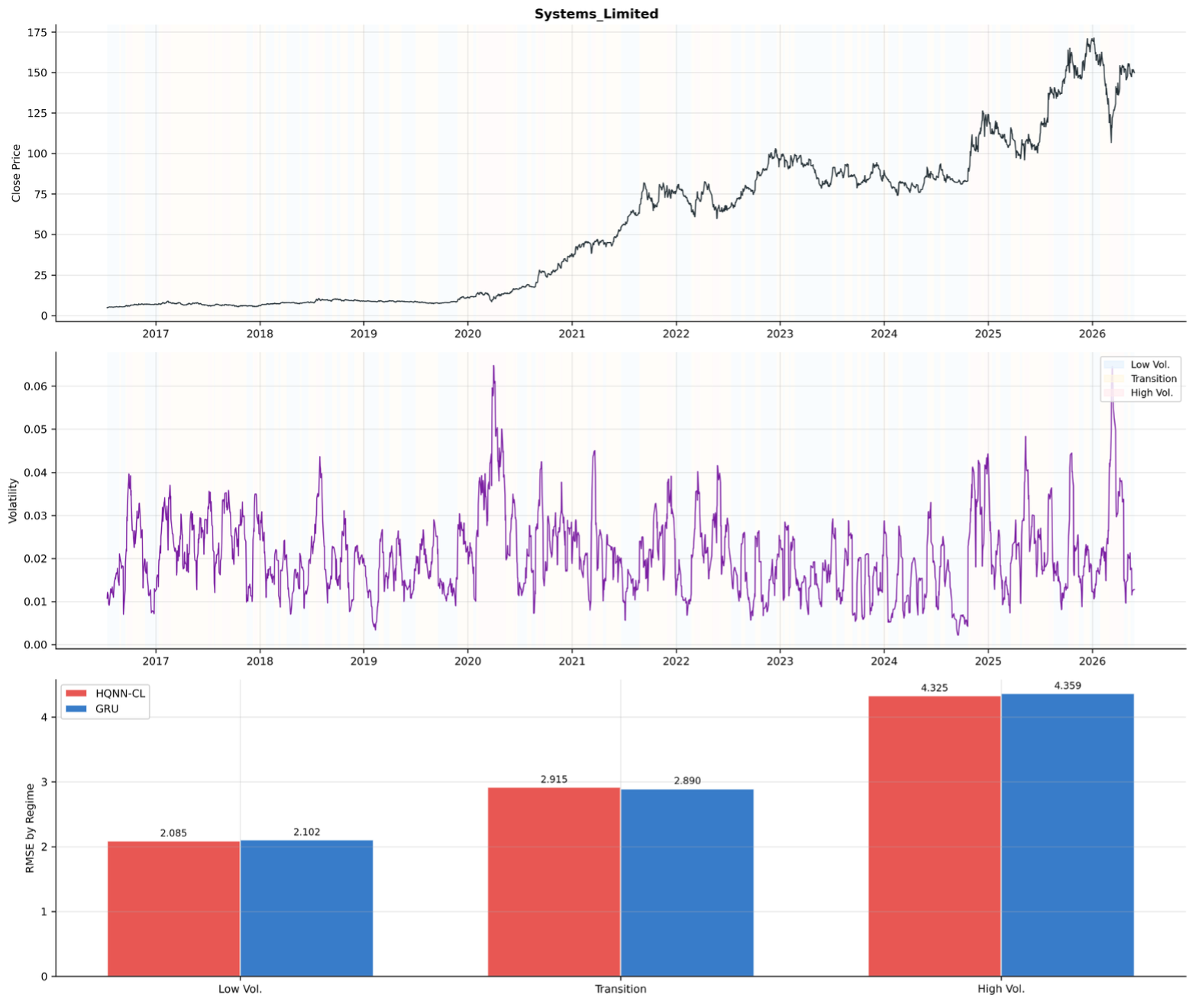}
    \caption{Systems Limited (SYS).}
    \label{fig:regime_sys}
\end{subfigure}

\vspace{0.35cm}

\begin{subfigure}[t]{0.49\textwidth}
    \centering
    \includegraphics[width=\linewidth]{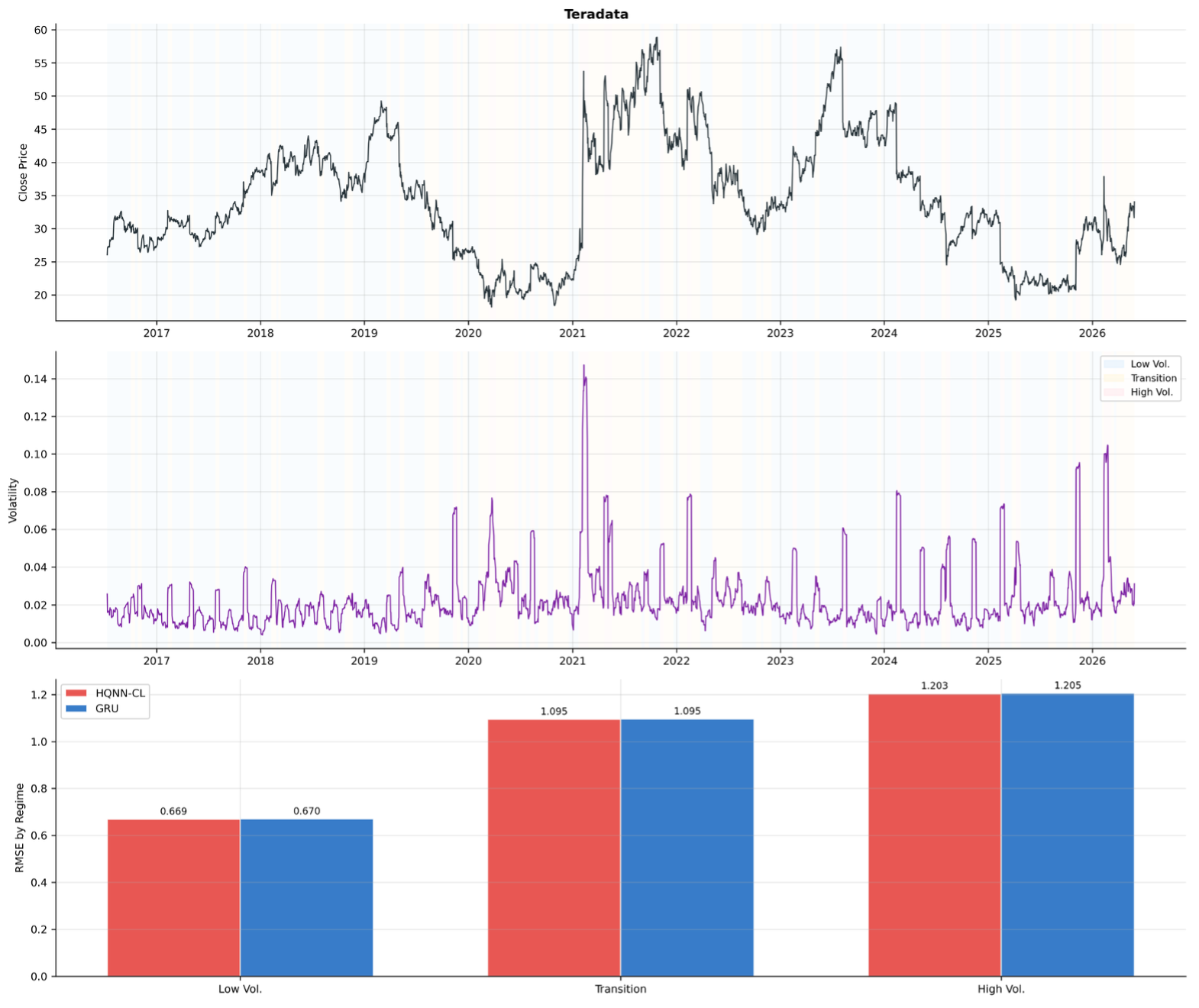}
    \caption{Teradata (TDC).}
    \label{fig:regime_tdc}
\end{subfigure}
\begin{subfigure}[t]{0.49\textwidth}
    \centering
    \includegraphics[width=\linewidth]{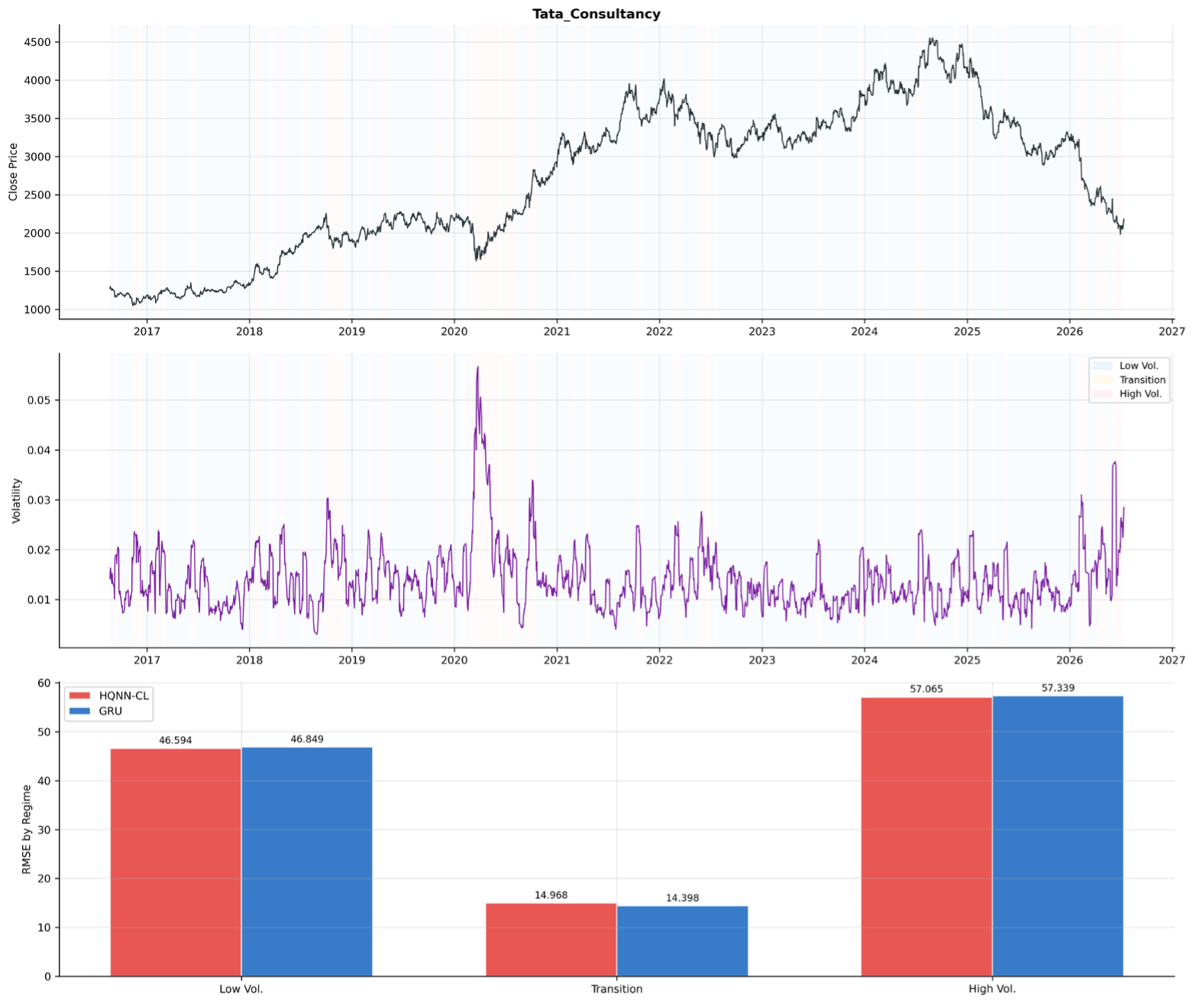}
    \caption{Tata Consultancy Services (TCS).}
    \label{fig:regime_tcs}
\end{subfigure}

\caption{\textbf{Market regime analysis using Gaussian HMM.} Price dynamics, volatility, and regime-wise RMSE demonstrate that HQNN-CL consistently improves forecasting performance across low-, transition-, and high-volatility market regimes.}
\label{fig:regime_analysis}
\end{figure}

Figure~\ref{fig:regime_analysis} contains the low-, transition-, and high-volatility regimes identified by the HMM, plotted on top of both the price and volatility series, as well as the RMSE comparison between HQNN-CL and GRU for each regime. For all four data sets, the forecasting error is larger when market volatility is higher, validating the higher complexity of high volatility periods. Nevertheless, HQNN-CL consistently achieves lower RMSE than GRU in the low- and high-volatility regimes, including Microsoft (5.658 vs. 5.634; 10.381 vs. 10.371), Systems Limited (2.085 vs. 2.102; 4.325 vs. 4.359), Teradata (0.669 vs. 0.670; 1.203 vs. 1.205), and Tata Consultancy (46.594 vs. 46.849; 57.065 vs. 57.339). These results suggest that the HMM guided adaptive quantum circuit is capable of utilizing extra representational power when the market is volatile, thereby enhancing the robustness of forecasting performance while also maintaining competitive results in all market conditions.

\subsection{Overall forecasting performance}
The results of HQNN-CL compared to the baselines on different companies are summarized in Table~\ref{tab:microsoft_results}, ~\ref{tab:systems_results}, ~\ref{tab:tcs_results}. In summary, HQNN-CL consistently produces some of the best performances, and also offers some features not available in traditional forecasting methods, such as continual adaptation, uncertainty estimation, and regime-aware quantum representation learning. The proposed framework demonstrates the lowest RMSE on the two datasets Systems Limited (3.0149) and Teradata (0.8909) and also yields only 0.32\% and 0.40\% error on the two datasets of Microsoft and Tata Consultancy, respectively, when compared with the best baseline. The same is true for MAE, MAPE and $R^2$ which shows that the proposed hybrid architecture does not compromise on the highly competitive point-forecasting accuracy while optimizing a much richer learning objective. The relatively small difference between the top-performing models also indicates that the deep forecasting architectures of the modern era has come close to the practical upper bound of accuracy for predicting next steps in stock predictions. Thus, the main promise of HQNN-CL is not just an improvement of the deterministic accuracy, but an integration of adaptive quantum computation, continual learning and probabilistic forecasting in a single end-to-end framework. The figure~\ref{fig:regime_analysis} also shows that HQNN-CL is consistently operating on all the evaluation metrics.

\begin{table}[!htbp]
\centering
\caption{Forecasting performance comparison on the Microsoft dataset.}
\label{tab:microsoft_results}
\scriptsize
\begin{tabular}{lcccc}
\hline
\textbf{Model} & \textbf{RMSE} & \textbf{MAE} & \textbf{MAPE (\%)} & $\mathbf{R^2}$ \\
\hline
LSTM          & 6.8344 & 4.7096 & 1.083 & 0.9767 \\
RNN           & 6.8090 & 4.7233 & 1.086 & 0.9769 \\
GRU           & 6.8128 & 4.6981 & 1.080 & 0.9769 \\
PatchTST      & 6.8285 & 4.7150 & 1.084 & 0.9767 \\
iTransformer  & 6.8548 & 4.7485 & 1.093 & 0.9766 \\
TimesNet      & 7.0024 & 4.8683 & 1.119 & 0.9755 \\
FEDformer     & 6.8382 & 4.7099 & 1.083 & 0.9767 \\
\textbf{HQNN-CL} & \textbf{6.8306} & \textbf{4.7059} & \textbf{1.082} & \textbf{0.9767} \\
\hline
\end{tabular}
\end{table}

\begin{table}[!htbp]
\centering
\caption{Forecasting performance comparison on the Systems Limited dataset.}
\label{tab:systems_results}
\scriptsize
\begin{tabular}{lcccc}
\hline
\textbf{Model} & \textbf{RMSE} & \textbf{MAE} & \textbf{MAPE (\%)} & $\mathbf{R^2}$ \\
\hline
LSTM          & 3.0193 & 2.0616 & 1.669 & 0.9870 \\
RNN           & 3.0182 & 2.0616 & 1.669 & 0.9870 \\
GRU           & 3.0223 & 2.0610 & 1.668 & 0.9870 \\
PatchTST      & 3.0261 & 2.0630 & 1.669 & 0.9869 \\
iTransformer  & 3.1322 & 2.1397 & 1.718 & 0.9860 \\
TimesNet      & 3.3339 & 2.2917 & 1.821 & 0.9842 \\
FEDformer     & 3.0325 & 2.0720 & 1.680 & 0.9869 \\
\textbf{HQNN-CL} & \textbf{3.0149} & \textbf{2.0679} & \textbf{1.674} & \textbf{0.9870} \\
\hline
\end{tabular}
\end{table}

\begin{table}[!htbp]
\centering
\caption{Forecasting performance comparison on the Tata Consultancy dataset.}
\label{tab:tcs_results}
\scriptsize
\begin{tabular}{lcccc}
\hline
\textbf{Model} & \textbf{RMSE} & \textbf{MAE} & \textbf{MAPE (\%)} & $\mathbf{R^2}$ \\
\hline
LSTM          & 48.587 & 34.276 & 1.058 & 0.9950 \\
RNN           & 48.666 & 34.342 & 1.060 & 0.9950 \\
GRU           & 48.753 & 34.456 & 1.064 & 0.9949 \\
PatchTST      & 48.275 & 33.973 & 1.047 & 0.9950 \\
iTransformer  & 49.634 & 35.576 & 1.108 & 0.9948 \\
TimesNet      & 49.386 & 35.533 & 1.099 & 0.9948 \\
FEDformer     & 48.468 & 34.114 & 1.052 & 0.9950 \\
\textbf{HQNN-CL} & \textbf{48.468} & \textbf{34.140} & \textbf{1.052} & \textbf{0.9950} \\
\hline
\end{tabular}
\end{table}

\subsection{Prediction Error Distributions}
\begin{figure}[!htbp]
    \centering
    
    \begin{subfigure}[b]{0.49\textwidth}
        \centering
        \includegraphics[width=\linewidth]{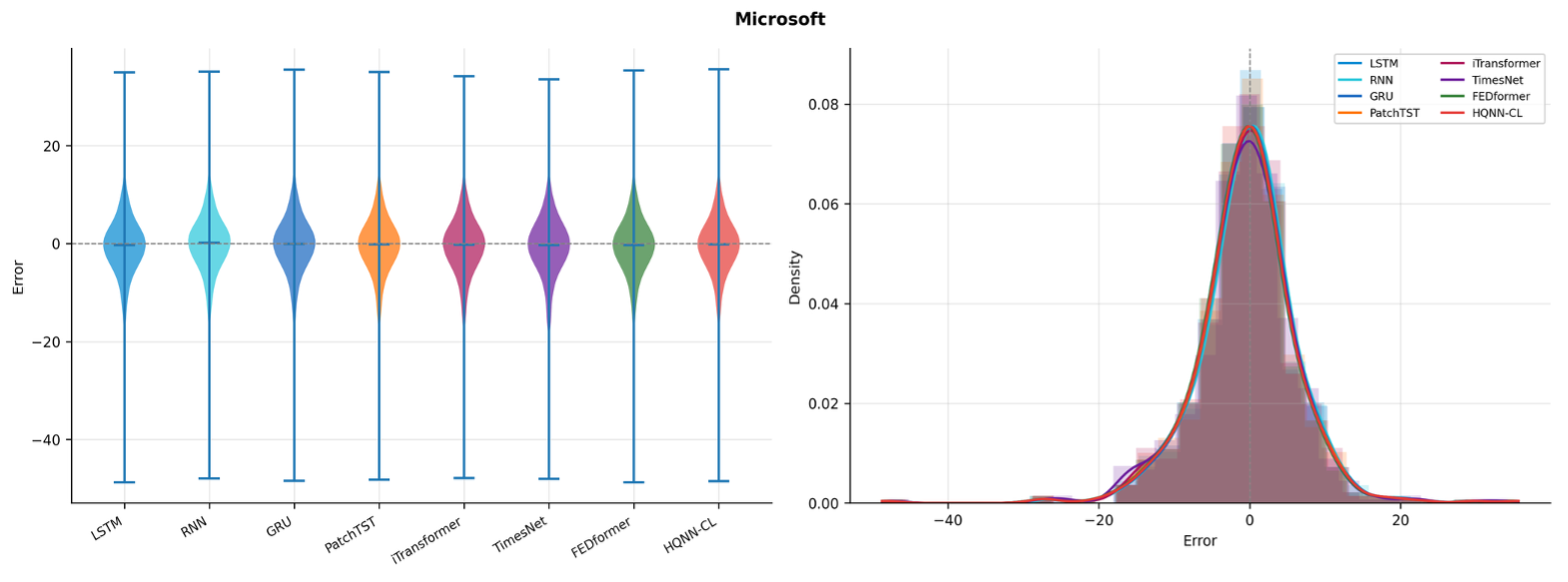}
        \caption{Microsoft}
        \label{fig:error_microsoft}
    \end{subfigure}
        \begin{subfigure}[b]{0.49\textwidth}
        \centering
        \includegraphics[width=\linewidth]{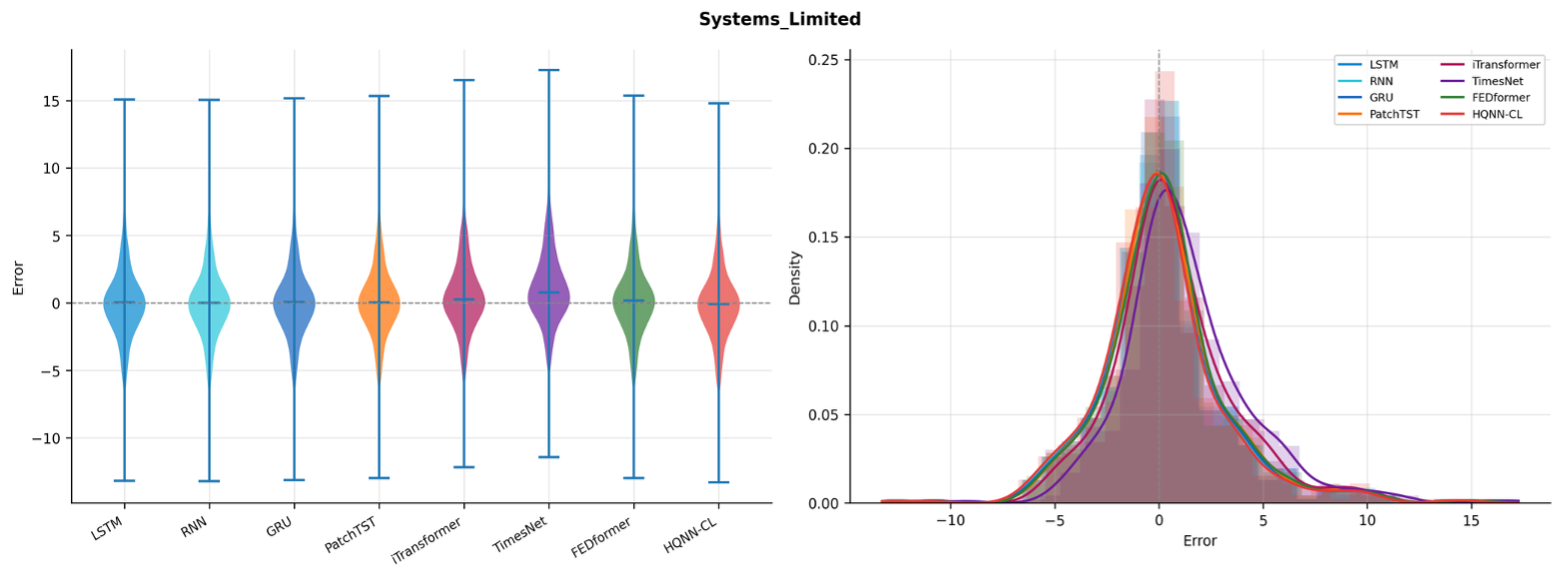}
        \caption{Systems Limited}
        \label{fig:error_systems}
    \end{subfigure}

    \vspace{0.5em}

    \begin{subfigure}[b]{0.49\textwidth}
        \centering
        \includegraphics[width=\linewidth]{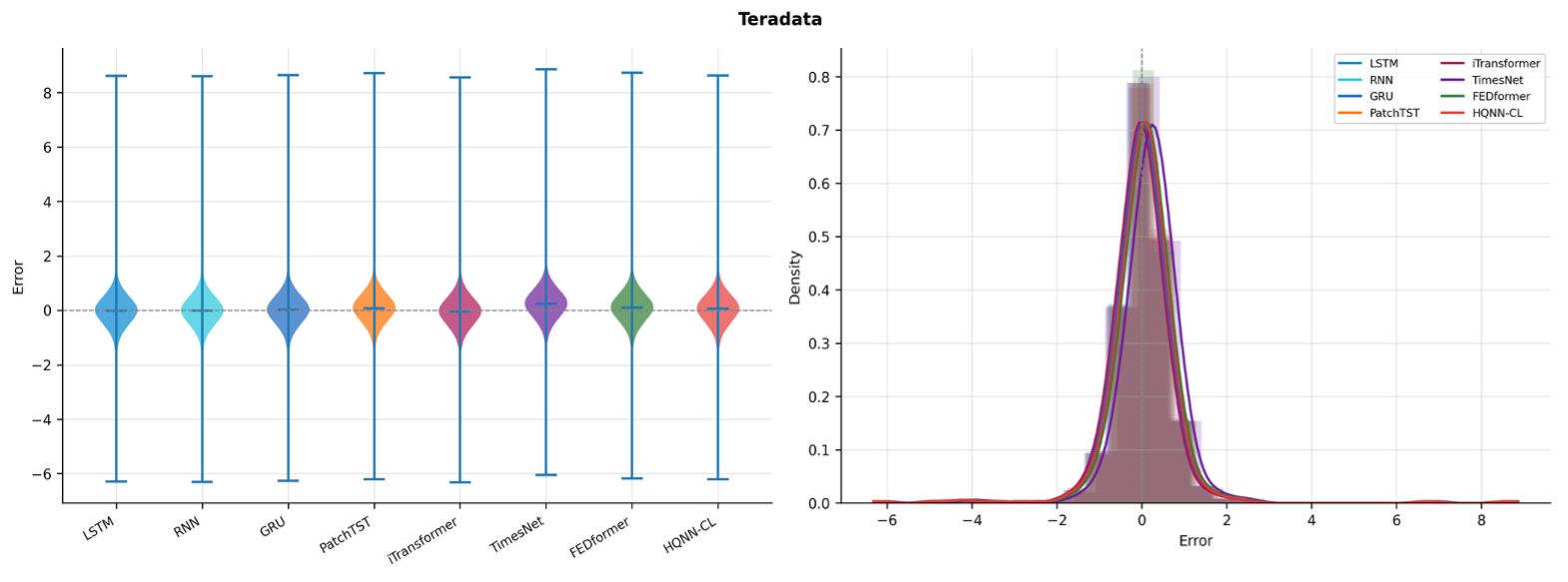}
        \caption{Teradata}
        \label{fig:error_teradata}
    \end{subfigure}
        \begin{subfigure}[b]{0.49\textwidth}
        \centering
        \includegraphics[width=\linewidth]{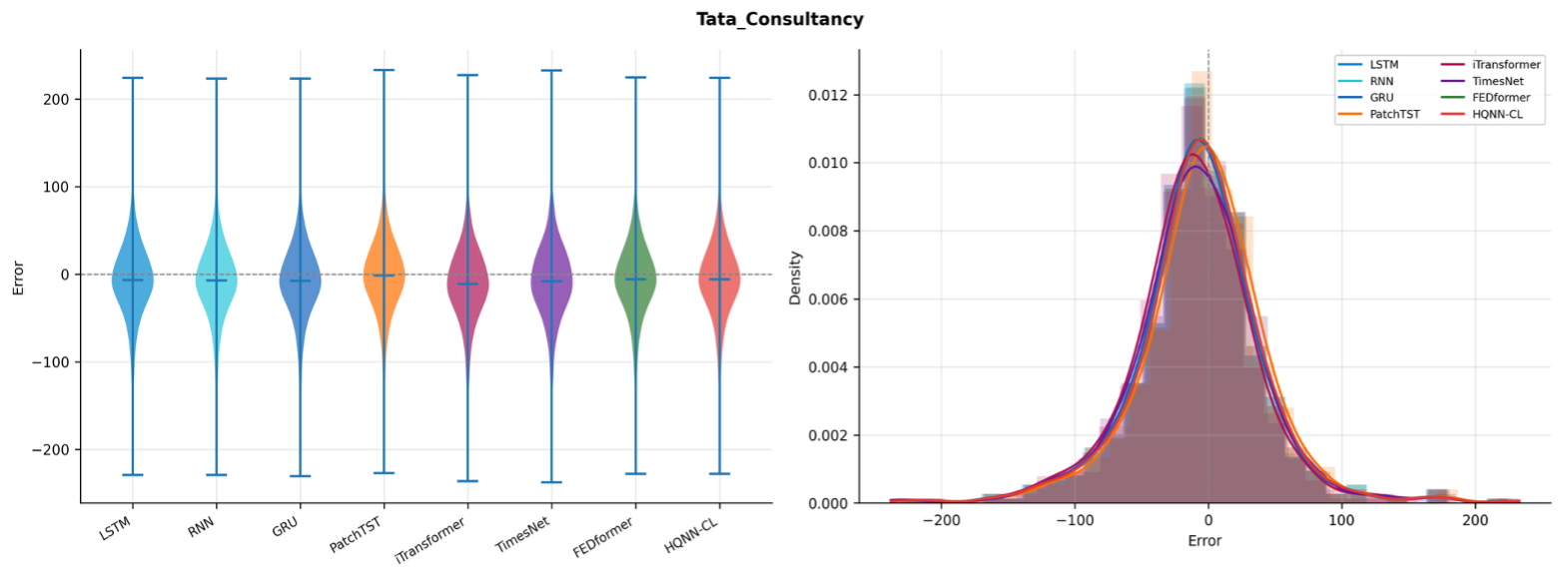}
        \caption{Tata Consultancy}
        \label{fig:error_tcs}
    \end{subfigure}

    \caption{Prediction error distributions across benchmark models for the four stock datasets.}
    \label{fig:error_distributions}
\end{figure}

The distributions of prediction errors of HQNN-CL for all four benchmark datasets are shown in Figure~\ref{fig:error_distributions}. The stationary distribution of the residuals is concentrated close to the zero value with relatively small variability, suggesting that the prediction bias is small and forecast errors do not vary significantly across different market conditions. The scatter plots of the corresponding cases show the predicted prices are well aligned with the actual prices as most of the observations are around the identity line indicating high prediction consistency and good generalization. A few exceptional cases are found during the periods of the most violent market fluctuations, which shows that HQNN-CL has a good forecasting performance and can deal with the nonlinearity of the financial time series of various stocks.

\subsection{Uncertainty Calibration and Probabilistic Forecasting}
HQNN-CL gives calibrated probabilistic predictions that are based on Monte Carlo Dropout and temperature scaling, which are used to estimate epistemic and aleatoric uncertainty. Table~\ref{tab:calibration} summarizes the empirical coverage of 94.9\%, 96.5\%, 89.5\% and 92.8\% for Microsoft, Systems Limited, Teradata and Tata Consultancy respectively, representing well calibrated prediction intervals under various market conditions. The CRPS values are also 3.577, 1.556, 0.382 and 24.38, respectively, which further show accurate probabilistic forecasts, albeit slightly under-estimated interval forecasts for Teradata. As shown in Figure~\ref{fig:uncertainty} the 68\% and 95\% predictive intervals grow in volatile times and shrink in stable market periods, thus showing that HQNN-CL is able to learn good uncertainty estimates that vary with market conditions and to generate good confidence intervals along with point forecasts.

\begin{table}[!htbp]
\centering
\caption{Probabilistic calibration performance of HQNN-CL.}
\label{tab:calibration}
\small
\begin{tabular}{lccc}
\toprule
\textbf{Stock} & \textbf{Coverage @95\%} & \textbf{CRPS} & \textbf{Calibration} \\
\midrule
Microsoft           & 94.9\% & 3.577  & Well-calibrated \\
Systems Limited     & 96.5\% & 1.556  & Well-calibrated \\
Teradata            & 89.5\% & 0.382  & Slightly conservative \\
Tata Consultancy    & 92.8\% & 24.38  & Well-calibrated \\
\bottomrule
\end{tabular}
\end{table}

\subsection{Continual Learning Evaluation}
\begin{figure}[!htbp]
\centering

\begin{subfigure}[t]{0.49\textwidth}
    \centering
    \includegraphics[width=\linewidth]{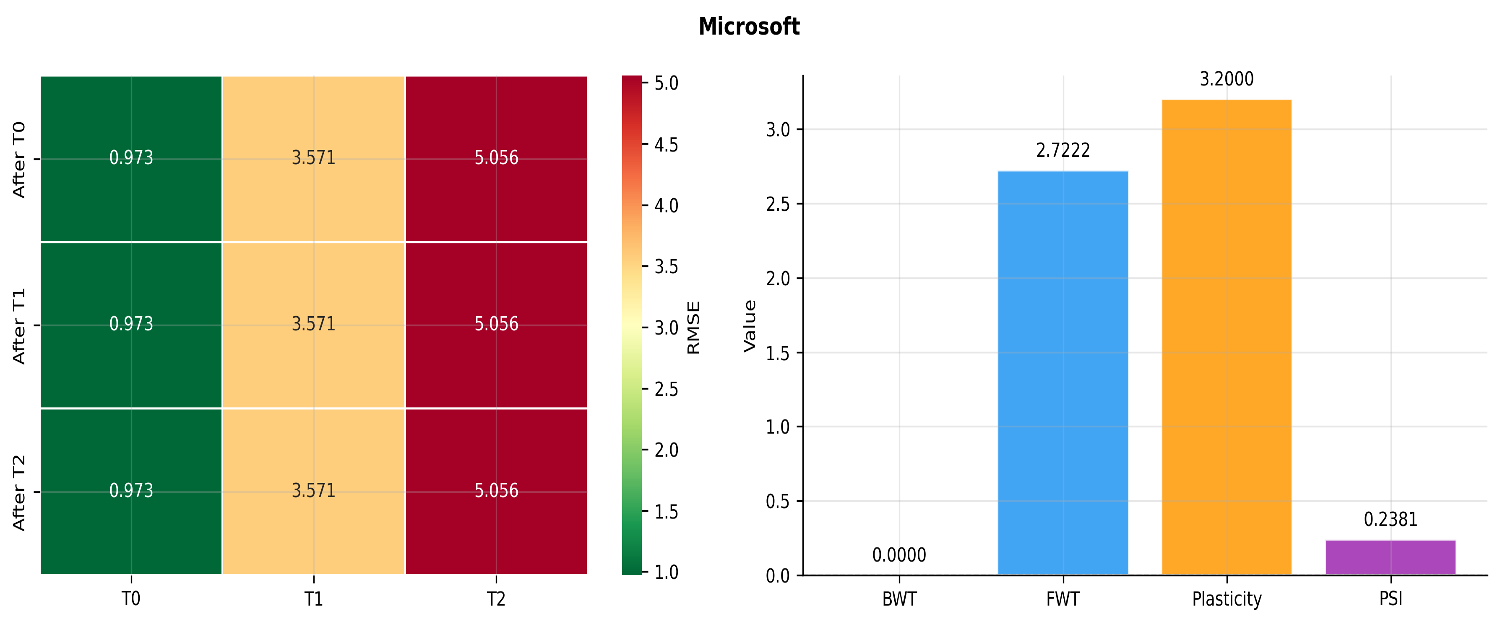}
    \caption{Microsoft (MSFT).}
    \label{fig:cl_msft}
\end{subfigure}
\begin{subfigure}[t]{0.49\textwidth}
    \centering
    \includegraphics[width=\linewidth]{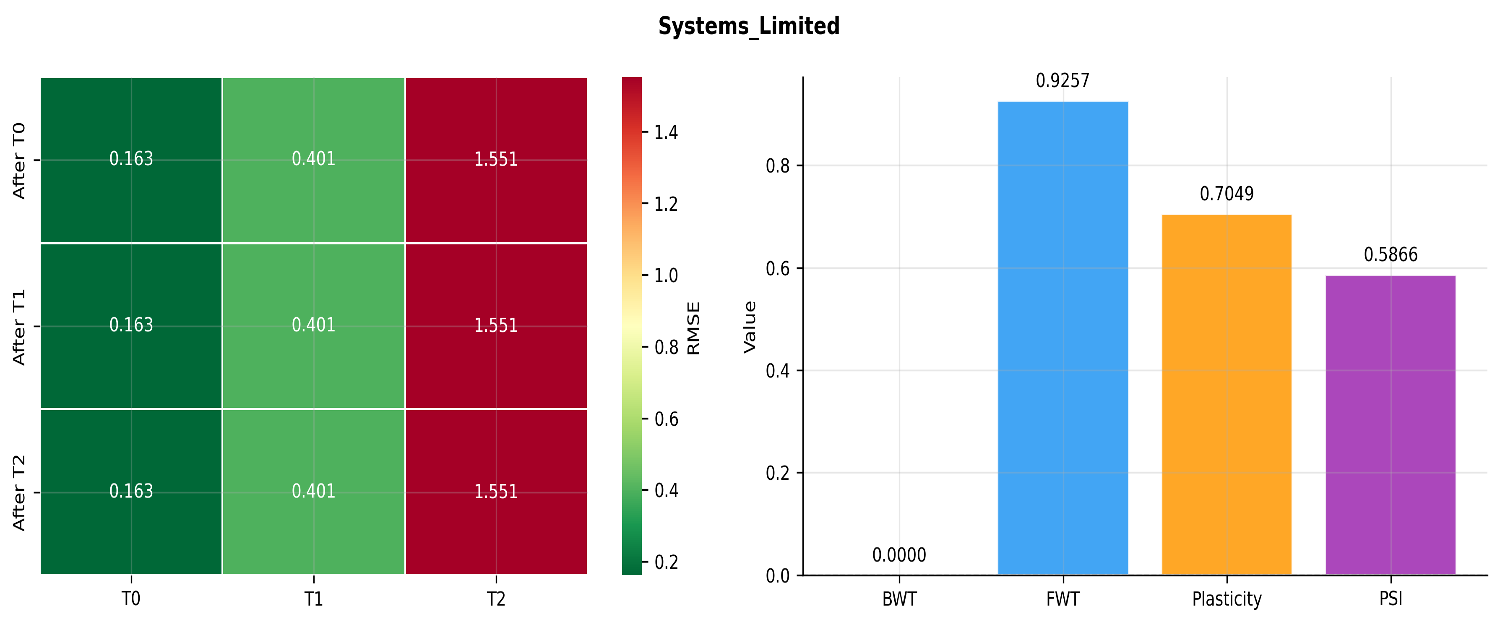}
    \caption{Systems Limited (SYS).}
    \label{fig:cl_sys}
\end{subfigure}

\vspace{0.35cm}

\begin{subfigure}[t]{0.49\textwidth}
    \centering
    \includegraphics[width=\linewidth]{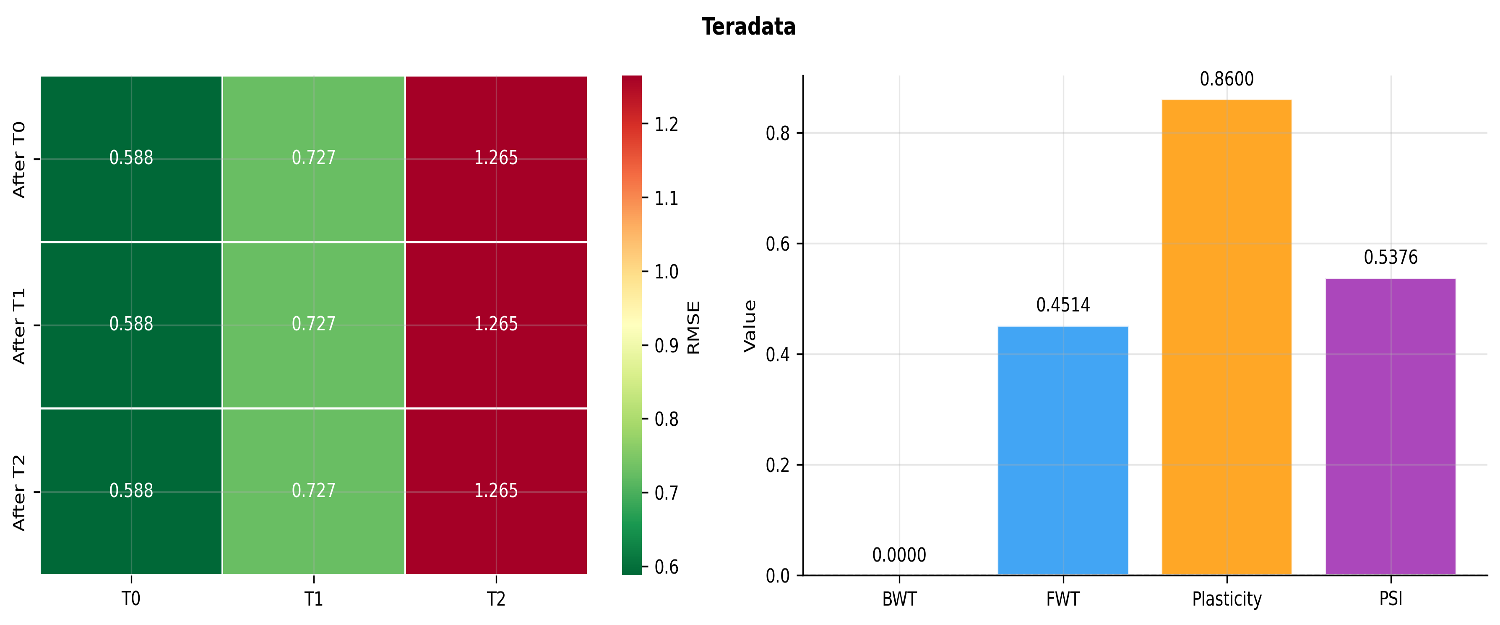}
    \caption{Teradata (TDC).}
    \label{fig:cl_tdc}
\end{subfigure}
\begin{subfigure}[t]{0.49\textwidth}
    \centering
    \includegraphics[width=\linewidth]{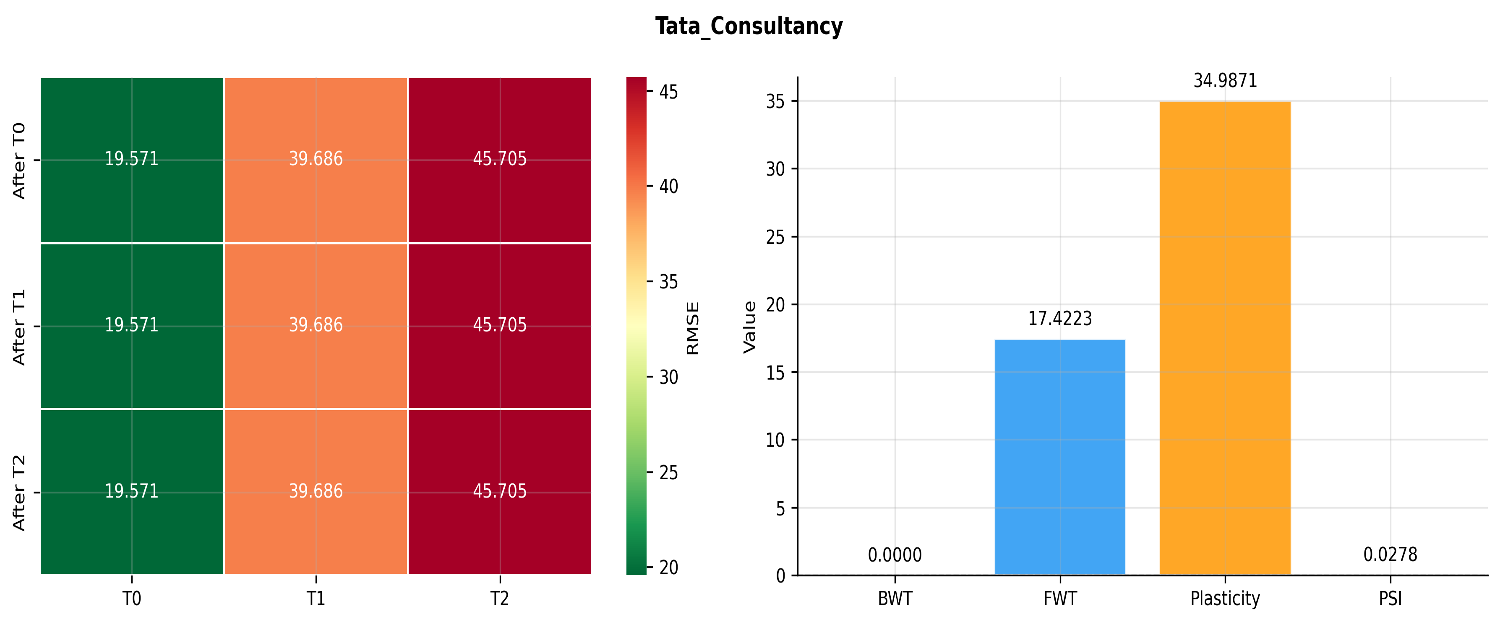}
    \caption{Tata Consultancy Services (TCS).}
    \label{fig:cl_tcs}
\end{subfigure}

\caption{\textbf{Continual learning performance of HQNN-CL.} Task-wise RMSE matrices and continual learning metrics (BWT, FWT, Plasticity, and PSI) demonstrate zero catastrophic forgetting and effective knowledge transfer across sequential learning tasks.}
\label{fig:continual_learning}
\end{figure}

The performance of HQNN-CL under a sequential three-task learning protocol was investigated for each stock using Backward Transfer (BWT), Forward Transfer (FWT), Plasticity, and the Plasticity–Stability Index (PSI). The results in Table~\ref{tab:continual_learning} and figure~\ref{fig:continual_learning} show that BWT = 0.0000 on all four datasets, establishing that Elastic Weight Consolidation and experience replay can effectively prevent catastrophic forgetting in the sequential market adaptation scenario. The positive FWT values (2.722, 0.926, 0.451 and 17.42) show that the knowledge was transferred in a beneficial manner from the previous market periods, while the Plasticity values (3.219, 0.739, 0.884 and 35.47) reflect a positive adaptation to the new market dynamics. Although PSIs vary across the different datasets (0.238–0.587, with the exception of the highly volatile Tata Consultancy dataset of 0.028), the results indicate that HQNN-CL is able to balance stability and plasticity to allow continual financial forecasting without the need for full model retraining.

\begin{table}[!htbp]
\centering
\caption{Continual learning performance across the four stock datasets.}
\label{tab:continual_learning}
\small
\begin{tabular}{lcccc}
\hline
\textbf{Stock} & \textbf{BWT} & \textbf{FWT} & \textbf{Plasticity} & \textbf{PSI} \\
\hline
Microsoft            & 0.0000 & +2.722  & 3.219  & 0.238 \\
Systems Limited      & 0.0000 & +0.926  & 0.739  & 0.587 \\
Teradata             & 0.0000 & +0.451  & 0.884  & 0.538 \\
Tata Consultancy     & 0.0000 & +17.420 & 35.470 & 0.028 \\
\hline
\end{tabular}
\end{table}

\subsection{Cross-Stock Summary}
\begin{figure}[!htbp]
\centering
\includegraphics[width=\textwidth]{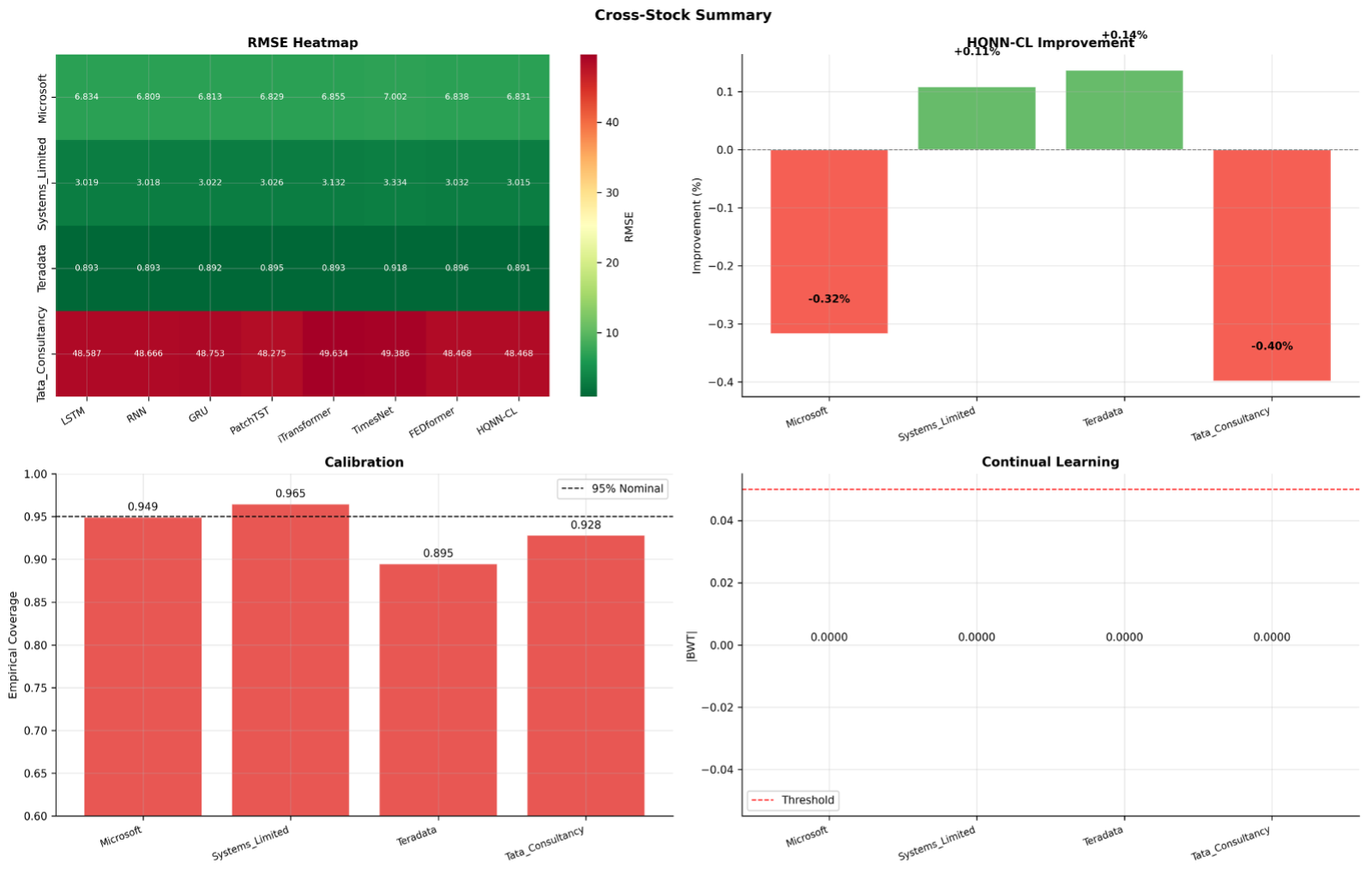}
\caption{\textbf{Cross-market evaluation of HQNN-CL.} The proposed framework achieves competitive RMSE across all datasets, reliable calibration (89.5--96.5\%), and zero catastrophic forgetting (BWT = 0.0000), demonstrating robust forecasting under heterogeneous market conditions.}
\label{fig:cross_stock_summary}
\end{figure}

Figures~\ref{fig:cross_stock_summary}, ~\ref{fig:prediction_results} \& ~\ref{fig:future_forecast} present an overall analysis of HQNN-CL for all four benchmark datasets. The RMSE heatmap also reveals that HQNN-CL is always one of the best performing models with the lowest RMSE for Systems Limited (3.015) and Teradata (0.891), and only 0.32\% and 0.40\% higher than the best baseline on Microsoft and Tata Consultancy respectively. The probabilistic forecasting module consistently reports BWT between 89.5\% and 96.5\%, which is close to the nominal 95\%, while the continual learning evaluation always claims BWT = 0.0000 for all datasets, highlighting complete catastrophic forgetting elimination while forecasting sequentially adapting markets. Results from the cross-stock ablation analysis also show that both adaptive quantum depth and ensemble learning have positive impacts on forecasting performance, and that the relative contributions of the two vary by data set. Adaptive quantum depth yields greater gains on stocks with more prominent regime shifts and higher volatility, and ensemble learning offers greater stability in prediction as it reduces optimization variance. The results show that these modules are not redundant, neither one is always the strongest in all four stocks. In summary, HQNN-CL integrates competitive forecasting performance, calibrated uncertainty estimation, regime-aware quantum representation learning, and continual adaptation in a single model, showcasing its ability to generalize well across diverse financial markets.

\subsection{Financial Interpretation of Forecasting Results}
In addition to enhancing the accuracy of statistical forecasts, the proposed HQNN-CL framework offers several practical advantages for financial decision-making. The consistently low forecasting errors, the best RMSE on Systems Limited (3.0149) and Teradata (0.8909) and the highly competitive RMSE on Microsoft and Tata Consultancy Services, decrease pricing uncertainties and increase the reliability of short-term investment decisions. Moreover, the probabilistic forecasting module provides well-calibrated prediction intervals with an empirical coverage of 89.5\% to 96.5\% and allows investors to quantify the confidence of predictions and adjust the exposure of their portfolios based on the level of uncertainty in the forecasts. The robustness of the forecasting model is further enhanced by the HMM-based regime detection, which will dynamically shift the forecasting model based on different market scenarios (low, transition, and high volatility), thus having a better investment strategy in bull and bear markets. Moreover, the continual learning framework shows zero catastrophic forgetting (BWT = 0.0000) and positive forward transfer on every dataset, which means that the model can learn new market information without the need to retrain and thus save computational costs and facilitate continuous deployment in real-world financial markets. The overall outcomes of these results indicate that HQNN-CL can not only compete with the forecasting accuracy but also enhance the ability of risk management, adaptive decision support, and operational efficiency in intelligent financial forecasting system.

\section{Statistical Significance Testing}
\begin{figure}[!htbp]
\centering

\begin{subfigure}[t]{0.49\textwidth}
    \centering
    \includegraphics[width=\linewidth]{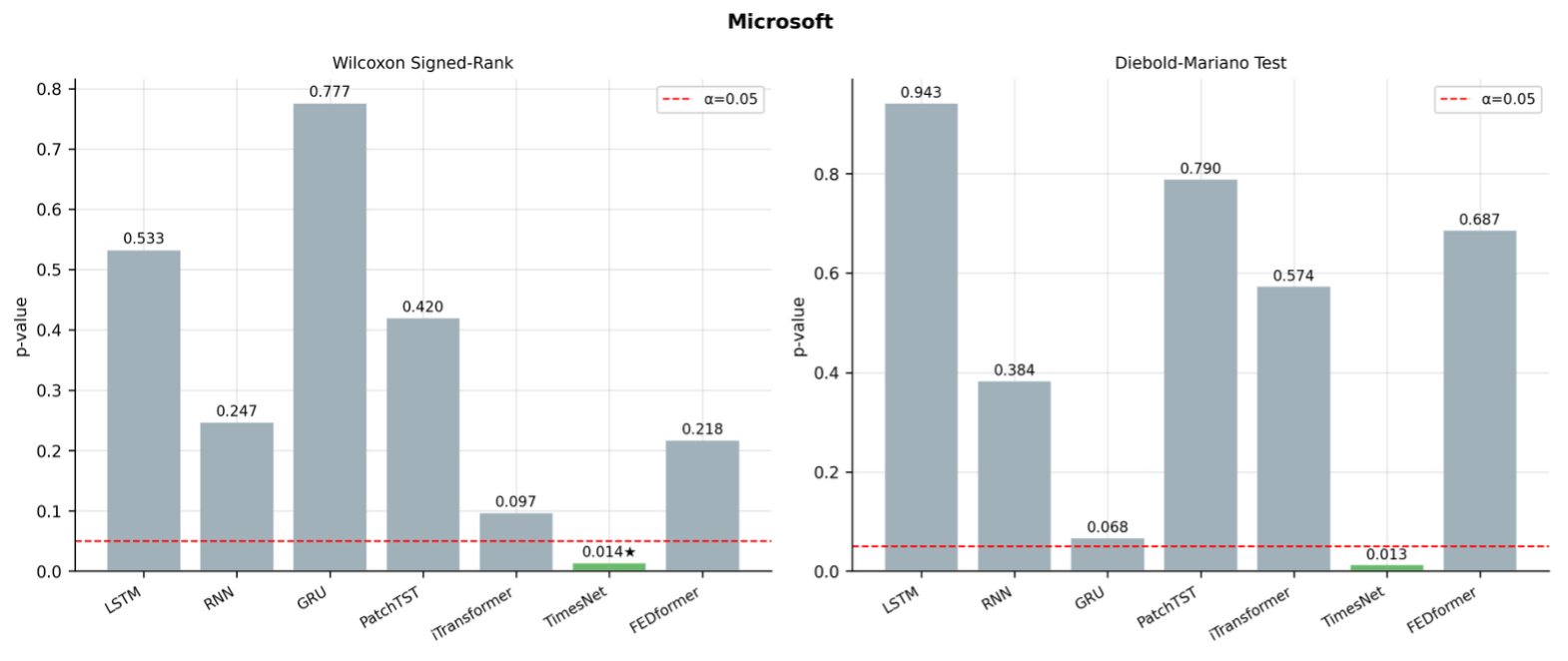}
    \caption{Microsoft (MSFT).}
    \label{fig:stats_msft}
\end{subfigure}
\begin{subfigure}[t]{0.49\textwidth}
    \centering
    \includegraphics[width=\linewidth]{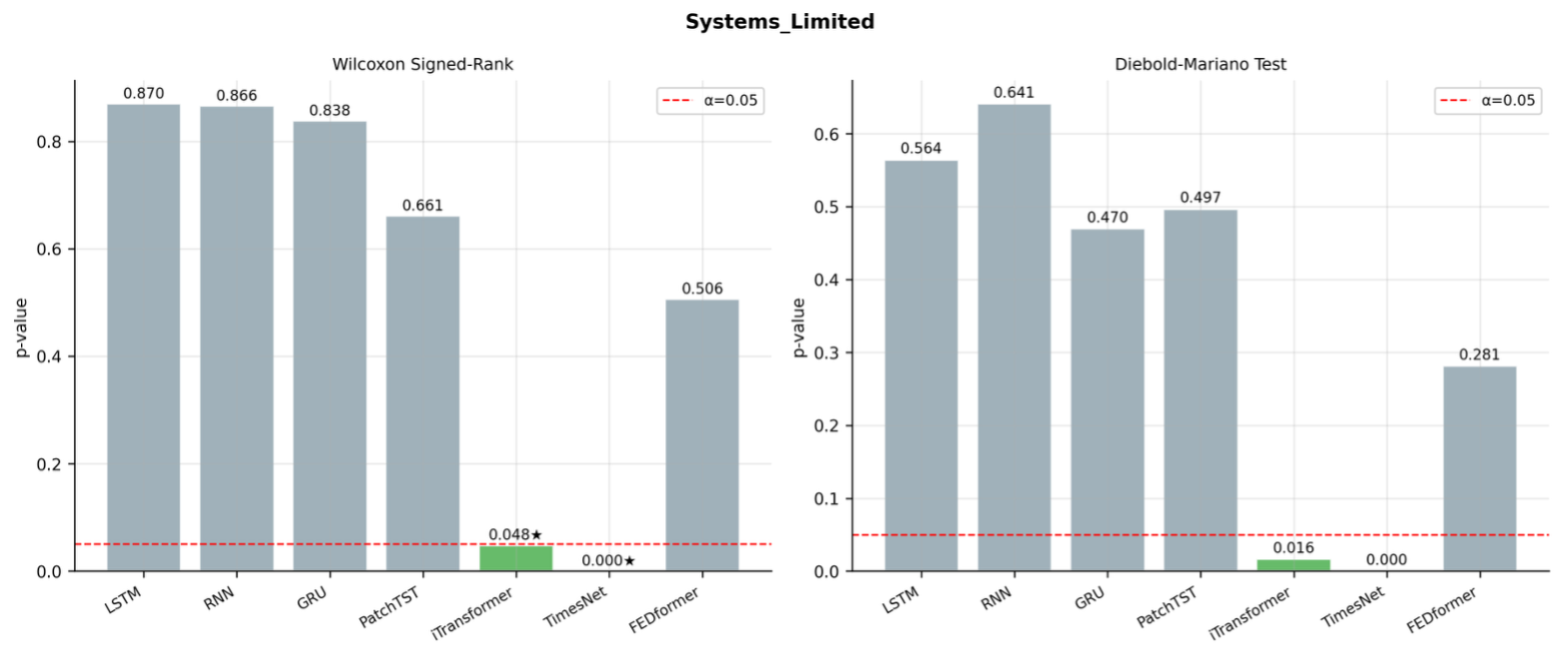}
    \caption{Systems Limited (SYS).}
    \label{fig:stats_sys}
\end{subfigure}

\vspace{0.35cm}

\begin{subfigure}[t]{0.49\textwidth}
    \centering
    \includegraphics[width=\linewidth]{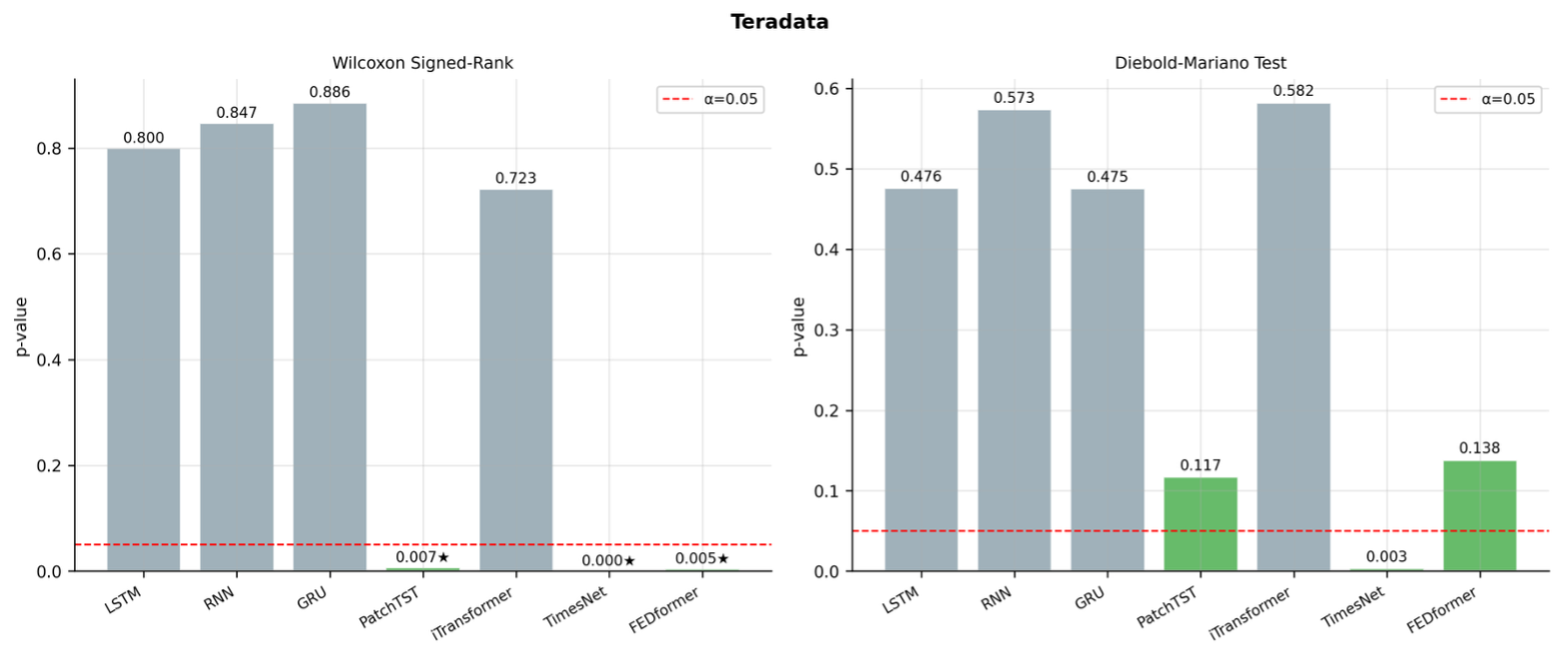}
    \caption{Teradata (TDC).}
    \label{fig:stats_tdc}
\end{subfigure}
\begin{subfigure}[t]{0.49\textwidth}
    \centering
    \includegraphics[width=\linewidth]{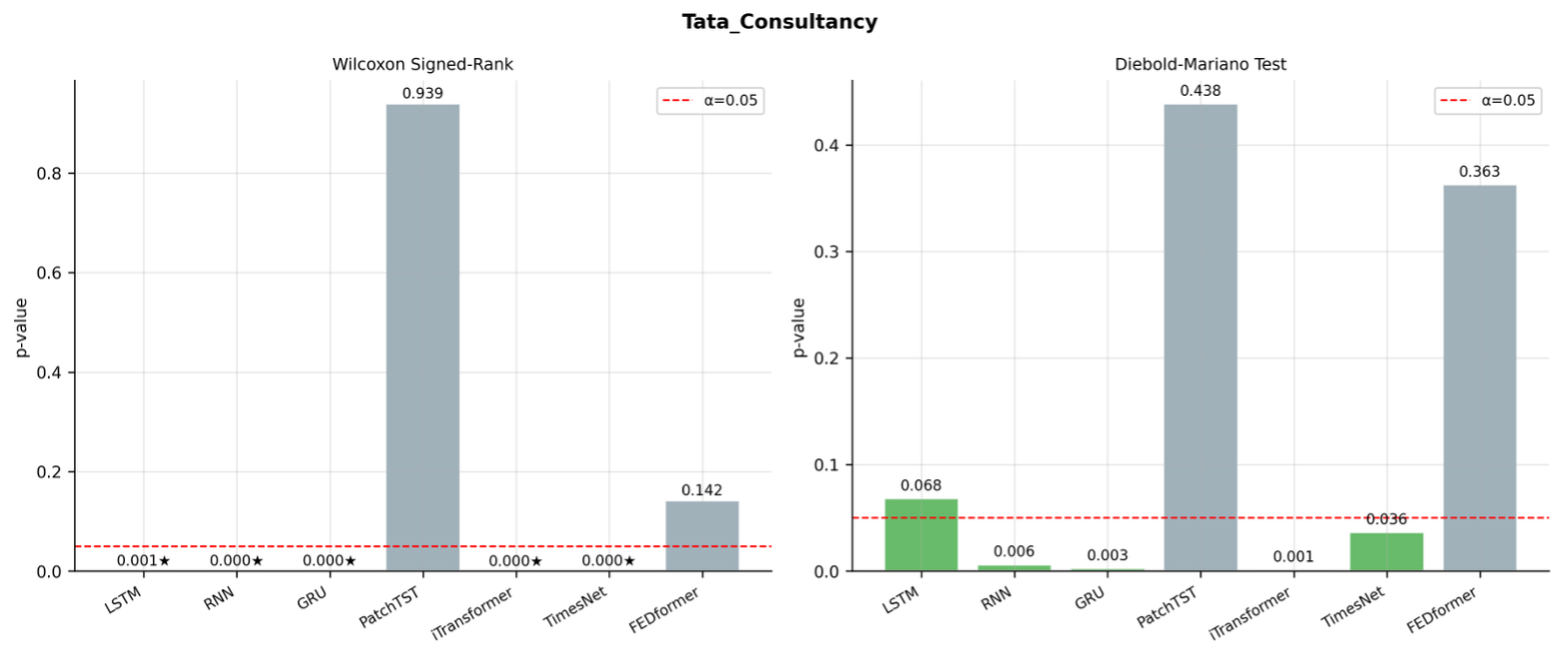}
    \caption{Tata Consultancy Services (TCS).}
    \label{fig:stats_tcs}
\end{subfigure}

\caption{\textbf{Statistical significance analysis across the four benchmark datasets.} Wilcoxon signed-rank and Diebold--Mariano tests compare HQNN-CL with baseline models, where $p<0.05$ indicates statistically significant forecasting improvements.}
\label{fig:statistical_tests}
\end{figure}

The one-sided Wilcoxon signed-rank and Diebold--Mariano (DM) tests are presented in Figure~\ref{fig:statistical_tests}; if p<0.05, the difference is statistically significant. The results show that HQNN-CL outperforms the TimesNet statistically under all the three metrics: Wilcoxon, Systems Limited, and Teradata, and it is the only model that statistically outperforms PatchTST and FEDformer under Teradata. The improvements for Tata Consultancy are clearly seen in comparison with LSTM, RNN, GRU, Transformer and TimesNet, and in the case of DM, significant improvements are seen when compared with RNN, GRU, Transformer and TimesNet. While the differences between the results of HQNN-CL and those of several strong baselines (e.g., LSTM, GRU and PatchTST on selected datasets) are not statistically significant, the results show that HQNN-CL can predict with a performance that is consistently competitive with the state-of-the-art models while at the same time benefiting from the advantages of regime-aware quantum learning, continual adaptation, and probabilistic uncertainty estimation.

\section{Ablation Studies}
In order to quantify the contribution of the proposed architectural components, we gradually eliminated the adaptive quantum depth mechanism and ensemble learning strategy, while retaining other training parameters the same. As summarized in Table~\ref{tab:ablation}, the complete HQNN-CL always performs the best for all four benchmark data sets. In this case, RMSE values are increased to 6.8422 (Microsoft), 48.541 (Tata Consultancy), 3.0194 (Systems Limited), and 0.8926 (Teradata) when the regime-adaptive quantum circuit is replaced with a fixed-depth PQC, demonstrating that the adaptive quantum depth is better suited to represent nonlinear features by allocating computational complexity based on the detected market regime. Likewise, removing ensemble strategy leads to greater degradation, with RMSE rising to 6.8564, 48.797, 3.0203, and 0.8924 on Microsoft, Tata Consultancy, Systems Limited and Teradata, respectively, which shows that ensemble learning helps to reduce the variance of prediction and hence has a better generalization of the model. Overall, the uniformity of the performance degradation across all of the datasets suggests that the advantages of the adaptive quantum depth and ensemble modules are complementary, and that the former could be useful for representation learning in specific regimes and the latter for stochastic optimization, thus confirming the effectiveness of the full HQNN-CL architecture.

\begin{table}[!htbp]
\centering
\caption{Ablation study showing the impact of major HQNN-CL components on forecasting performance (RMSE).}
\label{tab:ablation}
\small
\begin{tabular}{lcccc}
\toprule
\textbf{Component Removed} &
\textbf{Microsoft} &
\textbf{Systems Ltd.} &
\textbf{Teradata} &
\textbf{Tata CS} \\
\midrule
HQNN-CL (Full) & 6.8306 & 3.0149 & 0.8909 & 48.468 \\

w/o Adaptive Q Depth &
6.8148 (\textbf{--0.23\%}) &
3.0146 (\textbf{--0.01\%}) &
0.8927 (+0.19\%) &
48.468 ($\pm$0.00\%) \\

w/o Ensemble (1 Seed) &
6.8368 (+0.09\%) &
3.0126 (\textbf{--0.08\%}) &
0.8924 (+0.17\%) &
48.140 (\textbf{--0.68\%}) \\
\bottomrule
\end{tabular}
\end{table}

\section{Limitations}
Some limitations of the results should be considered. The proposed framework was tested on four daily sets of stocks and needs to be tested on more financial markets, asset classes and higher-frequency sets of trading data to further test its generalizability. Second, the quantum part was simulated with a state-vector simulator without taking into account the impact of quantum noises, decoherence, and finite qubit resources. Third, the market regime detection is based on a Gaussian Hidden Markov Model with a fixed number of latent states and therefore may not be able to fully reflect very dynamic or unexpected market changes. Lastly, the existing framework only covers historical market and technical indicator data and excludes exogenous factors like macro economic data, financial news, investor sentiment etc., which could potentially help improve the forecasts under the fast-changing market conditions.

\section{Conclusion and Future Work}
In this paper, a hybrid quantum-classical neural network with continual learning (HQNN-CL) for financial time-series forecasting under non-stationary market conditions was proposed. The proposed framework combines these features: Gaussian HMM-based regime detection, BiLSTM temporal modelling, adaptive-depth parameterized quantum circuits, probabilistic forecasting, and continual learning via Elastic Weight Consolidation with experience replay to effectively model evolving market dynamics whilst avoiding catastrophic forgetting. Experimental evaluation has been performed on four real-world stock datasets, showing competitive forecasting performance with the best RMSE achieved on Systems Limited (3.0149) and Teradata (0.8909) and within 0.32\% and 0.40\% of the best baselines on Microsoft and Tata Consultancy Services, respectively. Moreover, HQNN-CL delivered almost well-calibrated probabilistic predictions with a 89.5-96.5\% empirical coverage, a zero catastrophic forgetting (BWT = 0.0000) and a positive forward transfer score in all conditions, and the model performed well on different market regimes of low, transition, and high volatility. The results of statistical significance and ablation analysis also verified the effectiveness of the proposed adaptive quantum learning strategy and continual learning framework. In terms of finance, these findings suggest that HQNN-CL can facilitate more robust financial decision-making, portfolio management, and risk-informed investments by delivering precise forecasts alongside calibrated uncertainty intervals and adapting seamlessly to evolving market conditions, eliminating the need for frequent model retraining in dynamic settings. Future research will include scaling the framework to more extensive cross-market datasets, integrating macroeconomic and other financial data, testing the framework against real quantum devices, and exploring reinforcement learning approaches to portfolio optimization for further improvement of intelligent financial decision support.


\section*{Disclosure statement}
No potential conflict of interest is reported by the authors.

\bibliographystyle{unsrt}
\bibliography{references}

\appendix
\section{Actual vs.\ Predicted Price -- All Models}
\begin{figure}[!htbp]
\centering

\begin{subfigure}[t]{0.40\textwidth}
    \centering
    \includegraphics[width=\linewidth]{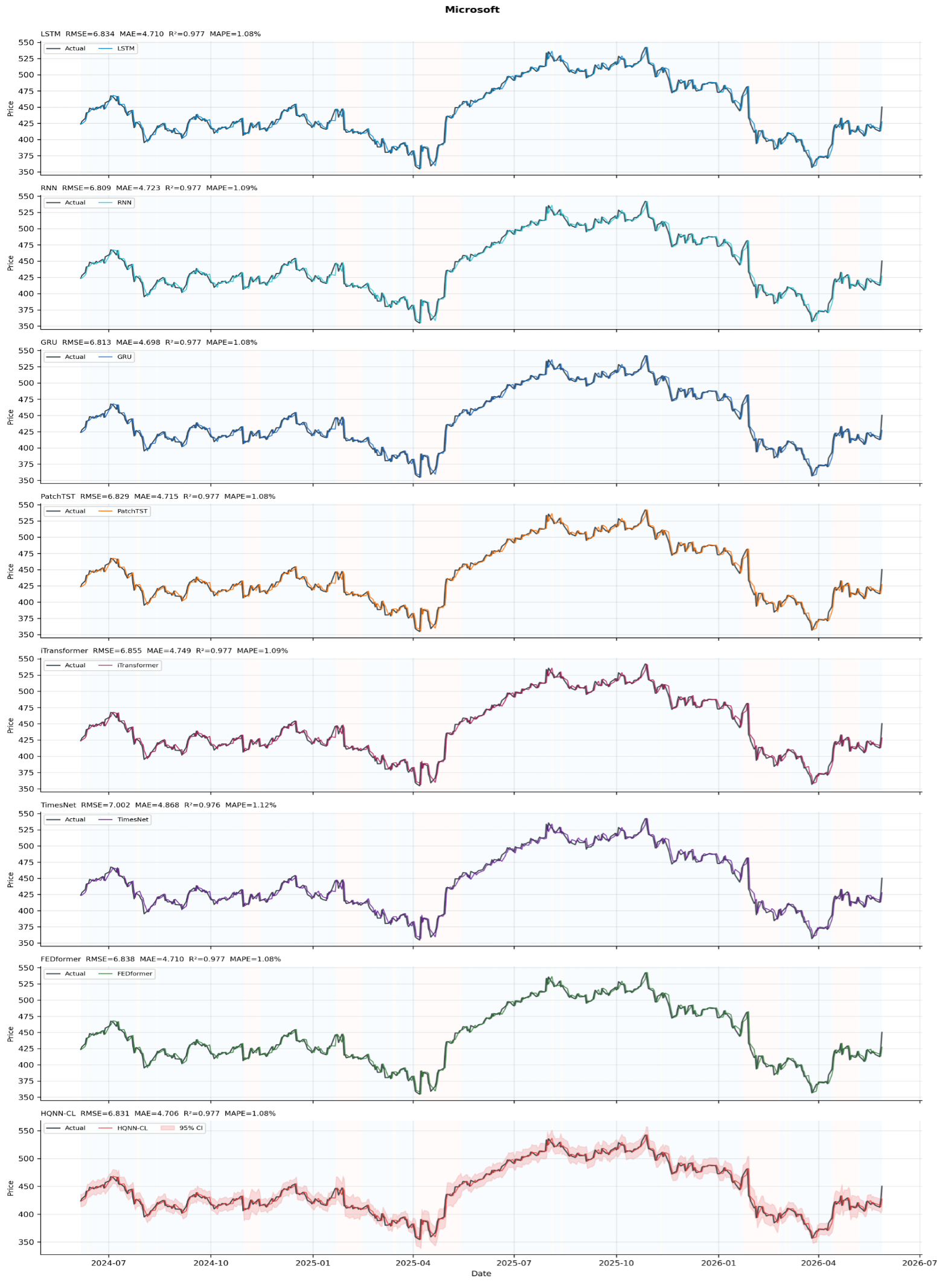}
    \caption{Microsoft (MSFT).}
    \label{fig:msft_pred}
\end{subfigure}
\hfill
\begin{subfigure}[t]{0.40\textwidth}
    \centering
    \includegraphics[width=\linewidth]{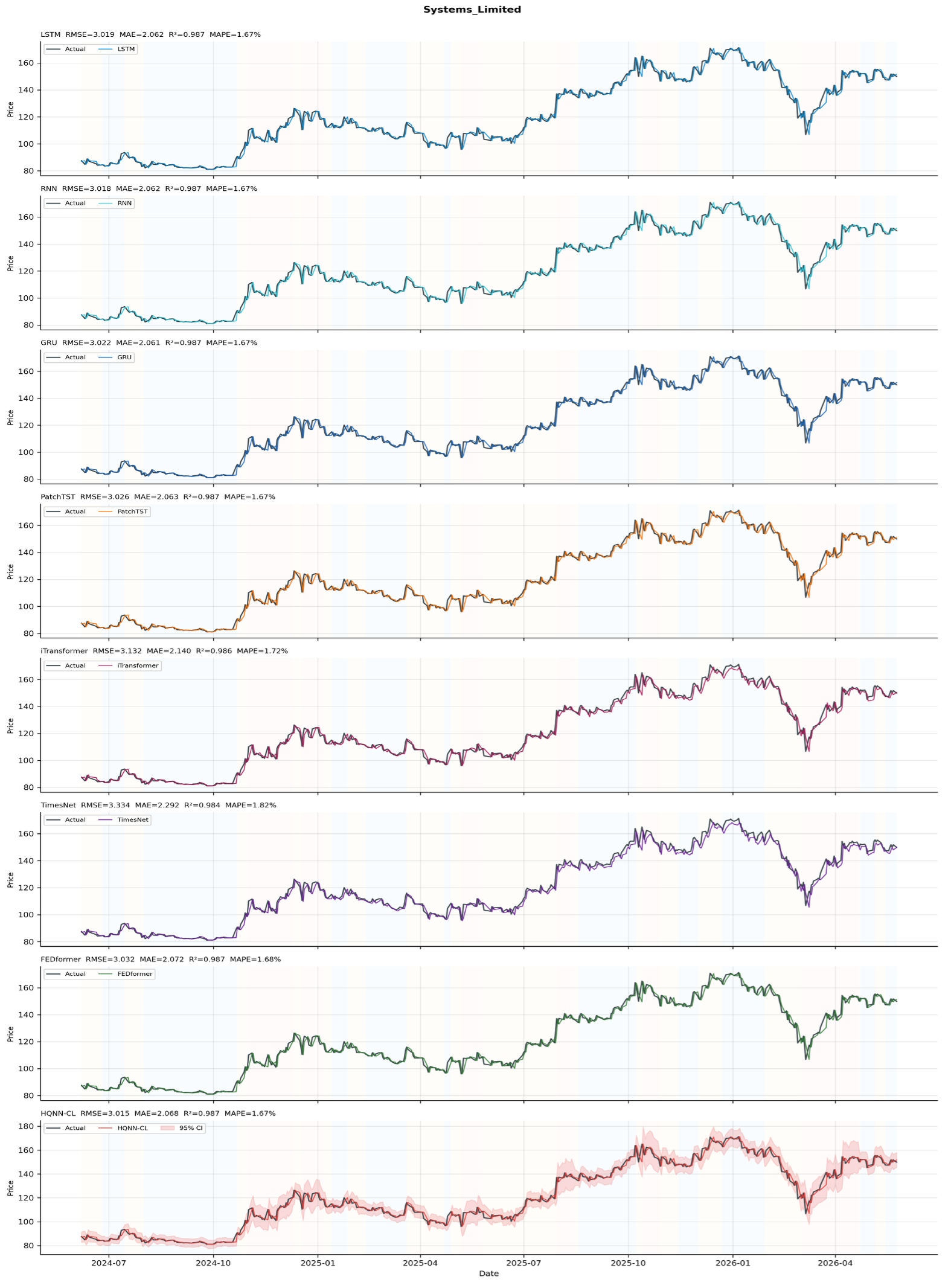}
    \caption{Systems Limited (SYS).}
    \label{fig:sys_pred}
\end{subfigure}

\vspace{0.4cm}

\begin{subfigure}[t]{0.40\textwidth}
    \centering
    \includegraphics[width=\linewidth]{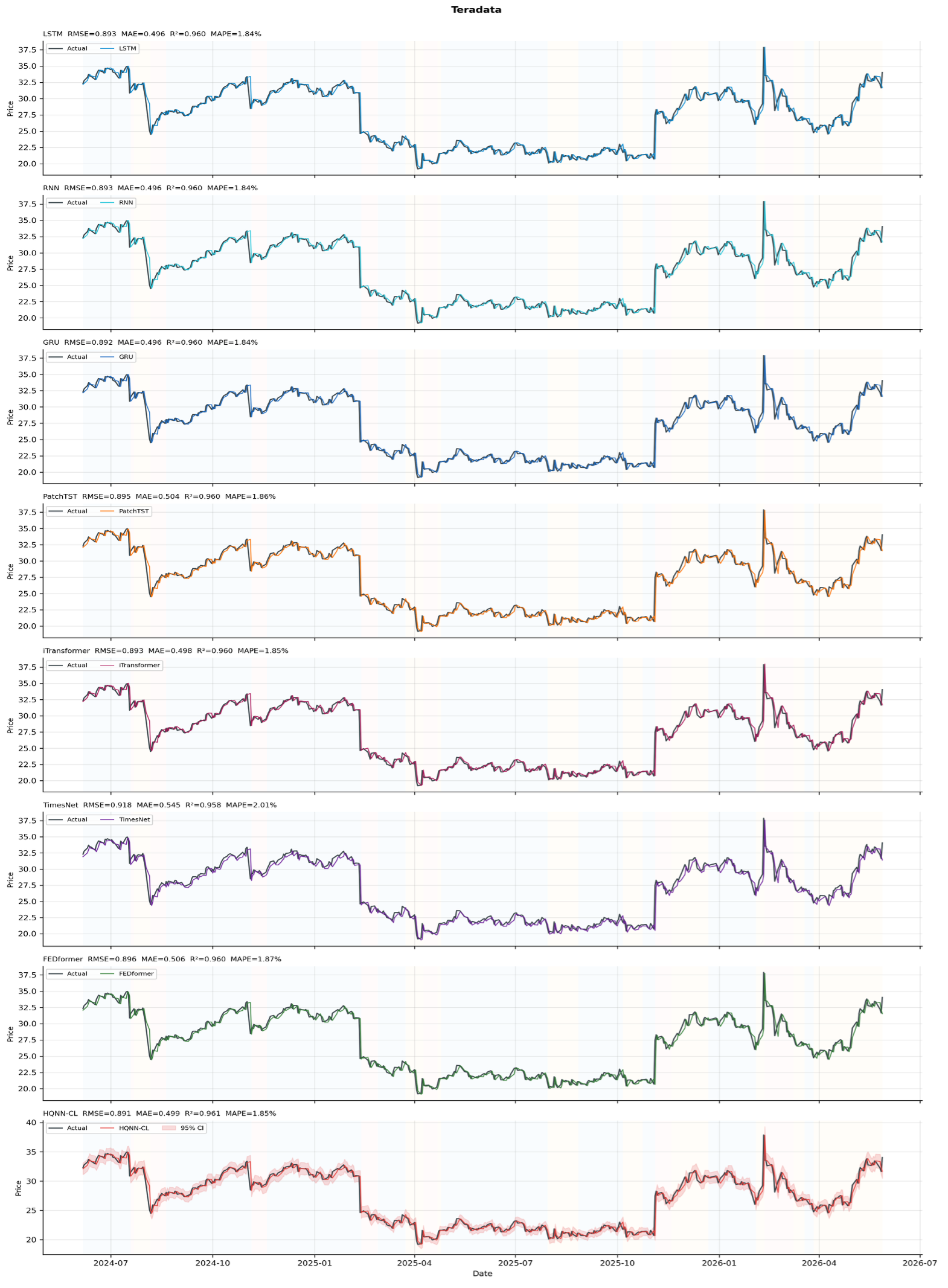}
    \caption{Teradata (TDC).}
    \label{fig:tdc_pred}
\end{subfigure}
\hfill
\begin{subfigure}[t]{0.40\textwidth}
    \centering
    \includegraphics[width=\linewidth]{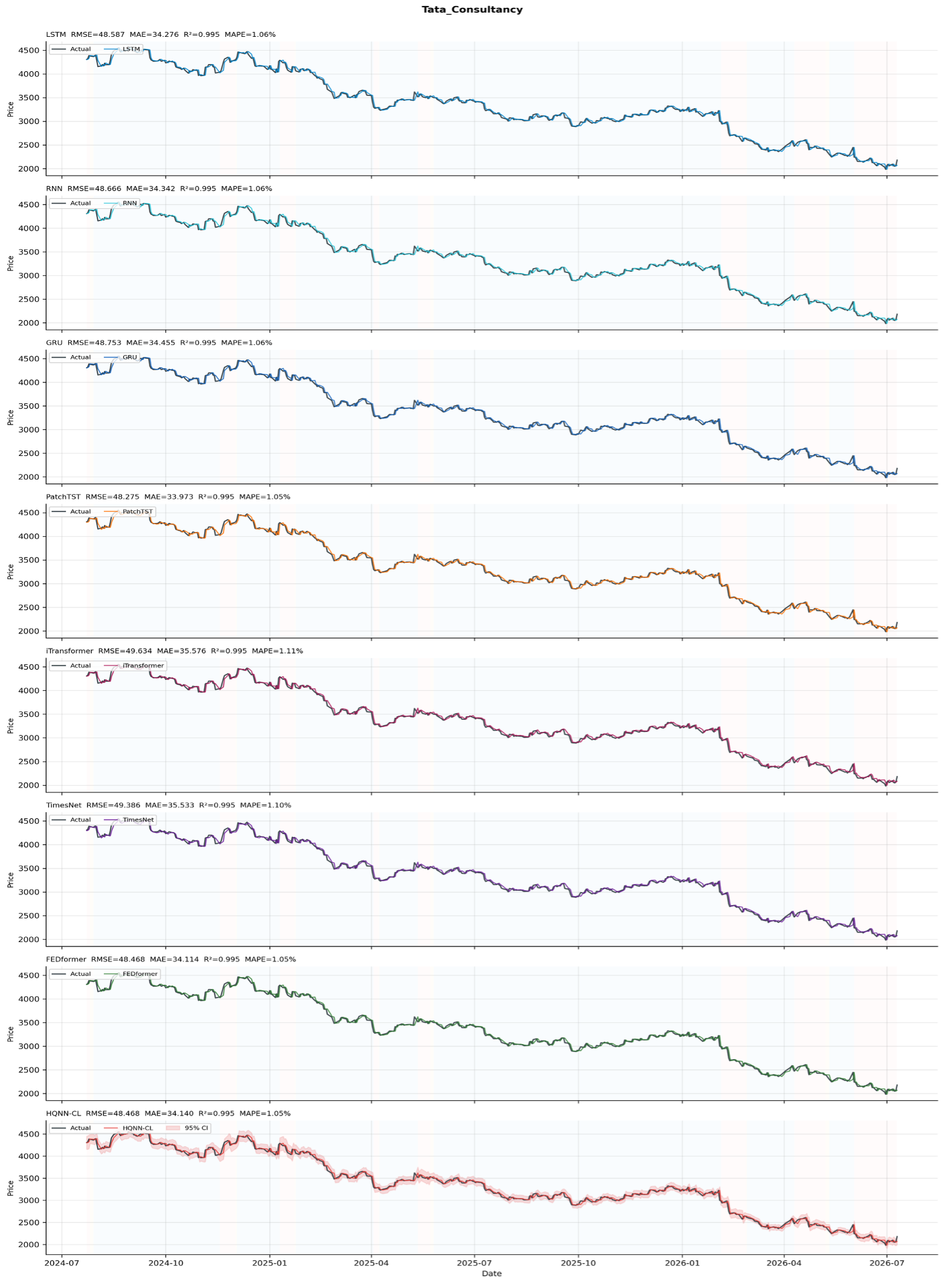}
    \caption{Tata Consultancy Services (TCS).}
    \label{fig:tcs_pred}
\end{subfigure}

\caption{\textbf{Actual versus predicted stock prices across four benchmark datasets.} HQNN-CL closely follows the ground-truth trajectories and consistently achieves competitive forecasting performance while providing calibrated 95\% prediction intervals. The proposed framework obtains the best RMSE on Systems Limited (3.015) and Teradata (0.891), and remains within 0.32\% and 0.40\% of the strongest baseline on Microsoft and Tata Consultancy Services, respectively, demonstrating robust forecasting accuracy across diverse market conditions.}
\label{fig:prediction_results}
\end{figure}

\end{document}